\documentclass{article}

\PassOptionsToPackage{numbers}{natbib}
\usepackage[main, preprint]{neurips_2026}

\usepackage[utf8]{inputenc} % allow utf-8 input
\usepackage[T1]{fontenc}    % use 8-bit T1 fonts
\usepackage{hyperref}       % hyperlinks
\usepackage{url}            % simple URL typesetting
\usepackage{booktabs}       % professional-quality tables
\usepackage{amsfonts}       % blackboard math symbols
\usepackage{nicefrac}       % compact symbols for 1/2, etc.
\usepackage{microtype}      % microtypography
\usepackage{xcolor}         % colors
\usepackage{graphicx}
\usepackage{amsmath}
\usepackage{cleveref}
\usepackage{booktabs}
\usepackage{multirow}
\usepackage{mathtools}

\newcommand{\bx}{\mathbf{x}}
\newcommand{\bxi}{\mathbf{x}^{(i)}}
\newcommand{\bz}{\mathbf{z}}
\newcommand{\ptz}{p_t\left(\mathbf{z}\right)}
\newcommand{\piz}{\pi\left(\mathbf{z}\right)}
\newcommand{\ptzx}{p_t\left(\mathbf{z} \mid \mathbf{x}\right)}
\newcommand{\ptzxi}{p_t\left(\mathbf{z} \mid \mathbf{x}^{(i)}\right)}
\newcommand{\score}{\nabla_\bz \log p_t\left(\bz\right)}
\newcommand{\pd}{p_\mathcal{D}}
\newcommand{\pdx}{\pd(\bx)}

\newcommand{\Qset}{\mathcal{Q}}
\newcommand{\Rset}{\mathcal{R}}
\newcommand{\Vset}{\mathcal{V}}

\title{Filtered Posterior Mean Collections: \\ A Unified Framework for Analytical Models of Diffusion Generalization}

\author{%
  Matthew Niedoba$^{1,2}$\\
  \texttt{mniedoba@cs.ubc.ca} \\
  \And
  Berend Zwartsenberg$^{2}$ \\
  \texttt{berend.zwartsenberg@inverted.ai}
  \And
  Frank Wood$^{1,2,3}$ \\
  \texttt{fwood@cs.ubc.ca}\\ [1em]
  \AFFS
  $^1$University of British Columbia, 
  $^2$Inverted AI, 
  $^3$Alberta Machine Intelligence Institute
}

\begin{document}

\maketitle

\begin{abstract}
      The neural-network denoising functions which form the backbone of image diffusion models are remarkably consistent in their generalization behaviour across a wide variety of network architectures and training procedure hyperparameters. A recent line of research has sought to model the outputs of these networks by aggregating posterior weighted averages of training dataset patches. In this work, we consolidate these approaches into a unified model class which we call Filtered Posterior Mean Collections (FPMCs). We define this model class using query precision vectors, response weights, and source distributions, and illustrate that existing methods are recoverable with specific choices of these design axes. Investigating each axis in turn, we find that FPMC performance can be improved with soft relaxations of prior patch-based methods, and through augmentations of source distributions. Applying these findings to an existing FPMC, we demonstrate consistent sample improvement across three natural image datasets.
\end{abstract}

\section{Introduction}
    \label{sec:intro}
% Paragraph 1 Diffusion models are great
Diffusion models \citep{sohl2015deep,ho2020denoising,song2020score} are a powerful class of generative models, especially in the domain of visual data where they have been widely adopted for image \citep{rombach2022high} and video \citep{harvey2022flexible} generation. These models generate samples using a neural-network parameterized denoising function which is trained to recover clean data from those corrupted by varying amounts of additive Gaussian noise. The unique minimizer of this denoising score matching objective \citep{vincent2011connection} is the optimal denoiser, a simple posterior weighted average over the elements of the training dataset \cite{karras2022elucidating}. However, this minimizer is a poor generative model that is incapable of generalization beyond the empirical training distribution \citep{gu2023memorization}.

% Paragraph 2 Image diffusion models generalize in similar ways
The generalization ability of diffusion models therefore implies that network denoisers learn biased approximations of the optimal denoiser \citep{yi2023generalization, song2025selective}. In the image domain, this bias is remarkably consistent across a wide variety of hyperparameters \citep{zhang2024emergence}. This observation has led to two complementary lines of research. The first of these seeks to understand the inductive biases responsible for denoiser convergence \citep{kadkhodaiegeneralization,niedoba2025towards,kamb2025analytic,bertrand2025closed,vastola2025generalization}, while the second aims to characterize the functional form of the resultant biased denoiser through network-free analytical approximations \citep{scarvelis2023closed,wang2024unreasonable,niedoba2024nearest,li2024understanding,kamb2025analytic,lukoianovlocality}.

% Paragraph 3 A collection of methods approximate diffusion model generalization in this specific way
Among analytical approaches, several methods have sought to approximate denoiser bias through posterior weighted averages of training dataset patches. These include \citet{niedoba2025towards} who propose patches based on average network gradients, \citet{kamb2025analytic} who utilize square patches and equivariant groups based on the inductive biases of convolutional networks, and \citet{lukoianovlocality} who derive patches from the eigendecomposition of the data covariance. Together, these methods represent the most accurate analytical approximations of network denoiser behaviour currently available. However, the design space of these methods lacks clarity, providing no clear directions for model improvement.

% Contribution 1: We unite these methods into a shared framework, and identify the primary design axes of the model class
In this work, we directly address this limitation by consolidating prior approaches into a unified model class which we call filtered posterior mean collections (FPMCs). Our framework defines FPMCs using sequences of query precisions $\Qset$ which filter the posterior means, response weights $\Rset$ which control aggregation, and source distributions $\Vset$ which vary the support of the averaging. Together, these three principal axes differentiate prior methodologies, and form the foundation for systematic improvement of model performance.

%defines three principal design axes which differentiate prior art, and elucidate the avenues for further model improvement.
We investigate the FPMC design axes in turn, in each case identifying and relaxing assumptions made by prior methodologies. Specifically, we improve FPMC denoisers by learning soft relaxations of prior patch-based methodologies, and through data augmentation strategies of their source distributions. When applied together, these modifications consistently improve the FPMC sample similarity versus diffusion model outputs on three natural image datasets. Our contributions are summarized as follows:

\begin{enumerate}
    \item We propose filtered posterior mean collections (FPMCs), a unified framework for analytical denoisers which approximate networks with aggregations of modified optimal denoisers. We elucidate $\Qset$, $\Rset$, and $\Vset$ as the principal design axes of this model class.
    \item We demonstrate that existing methods \citep{niedoba2025towards, kamb2025analytic, lukoianovlocality} are FPMCs recoverable through specific choices of $\Qset$,$\Rset$, and $\Vset$, clarifying the similarities and differences between prior approaches.
    \item We investigate the binary patch-based assumptions of prior methodologies, observing that FPMC estimation error can be reduced by independently or jointly finetuning $\Qset$ and $\Rset$ end-to-end.
    \item We study the effect of source distribution augmentation, finding that only horizontal reflection and synthetic augmentation consistently improve FPMC performance.
    \item We propose a general methodology for FPMC improvement based on joint finetuning of $\Qset$ and $\Rset$ along with data augmentation of $\Vset$. Applying this methodology to an existing FPMC, we achieve state of the art sample similarity improvement on CIFAR-10, FFHQ $64\times64$ and AFHQ $64\times64$.
\end{enumerate}
We will release our code publicly upon paper acceptance.

\section{Background}
    
 % 1. What is a diffusion process
The basis of diffusion models as well as many flow-based generative models is a stochastic forward process which gradually perturbs a base density $p(\bx), \bx \in \mathbb{R}^d$ with additive Gaussian noise. In this work, we consider $C$-channel image data of width $W $and height $H$, corresponding to $d=W\cdot H\cdot C$. Diffusion processes can be defined by a stochastic differential equation (SDE) using drift function $\mathbf{f}(\bz, t)$ and diffusion coefficient $g(t)$ of the form
\begin{equation}
    d\bz = \mathbf{f}(\bz, t) dt + g(t)d\mathbf{w} \label{eq:fwd_sde}.
\end{equation}
Integration of \cref{eq:fwd_sde} from initial values $(\bz,t) = (\bx, 0)$ to any $t \in (0,T]$ induces a conditional Gaussian distribution $\ptzx$. Through marginalization, one can obtain time-dependent distributions $\ptz = \int \ptzx p(\mathbf{x})d\mathbf{x}$ with $p_0(\bz) = p(\mathbf{x})$. Although, $\mathbf{f}$ can be an arbitrary function, it is convenient to select affine $\mathbf{f}(\bz, t) = f(t)\bz$, which results in $\ptzx = \mathcal{N}(\bz; \alpha(t), \sigma(t)^2\mathbf{I}_d) $ with closed form $\alpha(t)$ and $\sigma(t)$ \citep{lai2025principles}. Typically $f(t)$ and $g(t)$ are chosen such that $p_T(\bz) \approx \piz$, a tractable Gaussian prior distribution\footnote{Unless otherwise noted, we adopt the choices of \citep{karras2022elucidating} selecting $f(t) = 0$, $g(t) = \sqrt{2t}$, resulting in $\alpha(t)=1, \sigma(t)=t$.}.

% 2. PF-ODE and the score function
Diffusion models aim to learn a process which reverses \cref{eq:fwd_sde}. While this can be achieved through a corresponding reverse-time SDE, we will instead focus on the unique deterministic process 
\begin{equation}
    \frac{d\bz}{dt} = \mathbf{f}(\bz, t) - \frac{1}{2}g(t)^2\nabla_{\bz}\log p_t(\bz) \label{eq:pfode}
\end{equation}
whose marginal distributions also match those of \cref{eq:fwd_sde} for all $t \in (0,T]$. By leveraging this deterministic probability-flow differential equation (PF-ODE), diffusion models can generate samples by solving the initial value problem for $t=0$ with initial conditions $\bz \sim \pi(\bz)$ and $t=T$ via numerical integration techniques. 

Critically, integration of \cref{eq:pfode} relies on the score function $\nabla_\bz\log p_t(\bz)$. Diffusion models approximate this function using deep neural networks. One common method of approximating the score is through a denoising network $D_\theta$, optimized on the objective

\begin{equation}
    \mathop{\mathbb{E}}_{\substack{\bx, \bz \sim \ptzx p(\bx) \\ t\sim p(t)}} \left[\lambda(t) \left\lVert D_\theta(\bz, t) - \bx \right\lVert_2^2\right]. \label{eq:dsm}
\end{equation}
The denoiser is trained jointly over $t \in (0,T]$ distributed according to $p(t)$ and weighted via hyperparameter $\lambda(t)$. From this denoising model, the score can be estimated using Tweedie's formula

\begin{equation}
    \score = \frac{\alpha(t)\mathbb{E}\left[\bx \mid \bz, t\right] - \bz}{\sigma(t)^2} \approx \frac{\alpha(t)D_\theta(\bz, t) - \bz}{\sigma(t)^2} \label{eq:tweedies}
\end{equation}

% TODO: Segue sentence? 

% 3. Tweedie's, diffusion score matching

\subsection{Modelling Diffusion Generalization}
\begin{figure}[t!]
    \centering
    \includegraphics[width=0.9\linewidth]{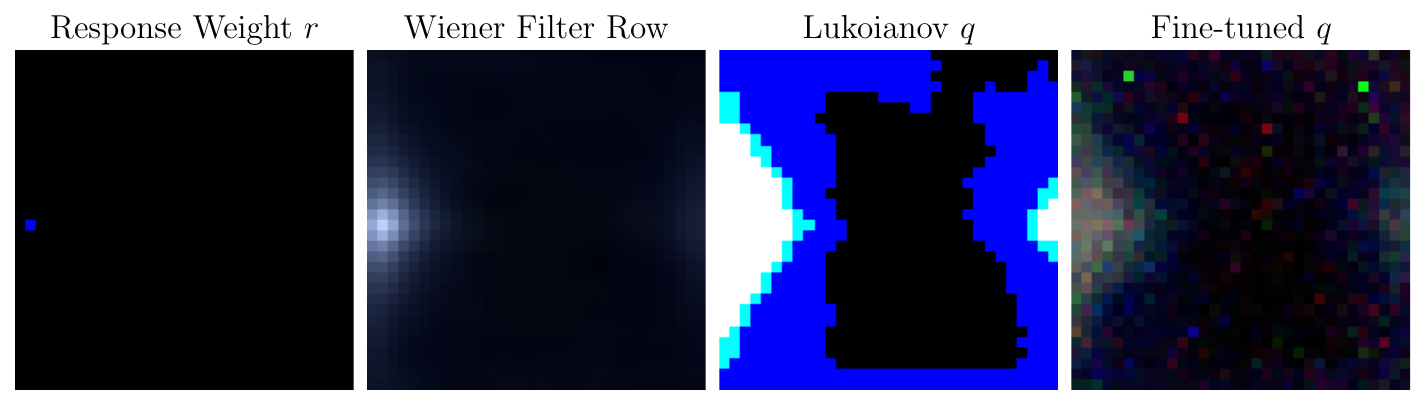}
    \caption{Visualization of $q$ and $r$ for the \citet{lukoianovlocality} FPMC on CIFAR-10 at $t=3.2$. All vectors are rescaled to $[0,1]$ for visualization. \textbf{Response weight $r$:} The one-hot response controls the output of the estimator, here a single channel of one pixel. \textbf{Wiener Filter:} The corresponding row of the Wiener filter exhibits a smooth, spatially localized structure. \textbf{Lukoianov $q$:} Thresholding destroys the graded structure of the Wiener filter. \textbf{Fine-tuned $q$:} Optimization recovers a soft structure which better matches the underlying Wiener filter.}
    \label{fig:q_viz}
\end{figure}

% Diffusion generalization consistency.
Remarkably, independent image diffusion models, when trained on the same data distribution, will often produce near-identical samples via \cref{eq:pfode} when identically initialized from the same $\bz$. This behaviour is robust across model architectures, training procedures, and numerical samplers \citep{zhang2024emergence}, and even when models are trained on disjoint subsets of the same training dataset \cite{kadkhodaiegeneralization}. Furthermore, this consistency extends beyond samples to the denoiser functions themselves \citep{niedoba2025towards}. Together, these findings suggest that generalization in diffusion models is the result of a consistent mapping from $\piz$ parameterized by the generalization behaviour of denoiser neural-networks.

% NEED segue about why we wish to approximate the behaviour of neural networks.
% The optimal denoiser
To gain insight into this generalization behaviour, analytical denoisers have been proposed to approximate black-box neural-network denoisers. One such analytical denoiser can be derived from \cref{eq:tweedies} which shows that the score is related to the posterior mean $\mathbb{E}\left[\bx \mid \bz, t\right]$. For general $p(\bx)$, this expectation is intractable. However, under a finite training set $\mathcal{D} = \{\bx^{(1)}, \ldots, \bx^{(N)} \mid \bx^{(i)} \sim p(\bx)\}$ and corresponding empirical data distribution $p_{\mathcal{D}}(\bx) = \frac{1}{N}\sum_{\mathbf{x}^{(i)} \in \mathcal{D}}\delta(\bx - \bx^{(i)})$, the empirical posterior mean has the closed form
\begin{align}
    \mathbb{E}_{p_\mathcal{D}}\left[\bx \mid \bz, t\right] &= \int p_t(\bx \mid \bz) \ \bx \ d\bx =  \sum_{\bx^{(i)} \in \mathcal{D}} p_t(\bx^{(i)} \mid \bz) \ \bx^{(i)},  \label{eq:emp_pmean}\\
    p_t(\bx \mid \bz) &= \frac{\ptzx p_\mathcal{D}(\bx)}{\sum_{\bx^{(i)} \in \mathcal{D}}\ptzxi p_\mathcal{D}(\bx^{(i)})}. \label{eq:posterior}
\end{align}
Although this \emph{optimal} denoiser described by \cref{eq:emp_pmean} is the minimizer of \cref{eq:dsm} \citep{vincent2011connection,karras2022elucidating}, it is a poor approximator of the behaviour of diffusion models for two reasons. First, its estimates do not match those of neural-network denoisers \citep{niedoba2025towards}, and  secondly, because the optimal denoiser is incapable of producing samples outside of the training dataset \citep{gu2023memorization}. To address these shortcomings, a wide variety of methods propose modifications to the optimal denoiser to better approximate network behaviour \citep{scarvelis2023closed,niedoba2025towards,kamb2025analytic,lukoianovlocality}.
% Optimal denoiser modifications

\section{Filtered Posterior Mean Collections}
    \begin{table}[t!]
    \caption{Existing methodologies can be reframed as FPMCs under specific choices of $\Qset$, $\Rset$ and $\Vset$. A more detailed summary of each methodology is provided in \cref{ap:prior_methods}}
    \centering
    {
        \renewcommand{\arraystretch}{1.3} 
        \resizebox{\textwidth}{!}{
            % \begin{tabular}{lllll}
% \hline
% \textbf{Denoiser} & $L$  & $\mathcal{Q}$ & $\mathcal{R}$ & $\mathcal{V}$ \\
% \hline
% Optimal & 1 & Identity & Identity & $\pdx$ \\
% LS \citep{kamb2025analytic} & $H^2$ & Binary Square Patches & Per-Pixel  & $\pd$ \\
% ELS \citep{kamb2025analytic} & $H^2$ & Binary Square Patches & Per-Pixel & $p_{\Omega(\mathcal{D})}(\bx)$ \\
% PSPC-Square \citep{niedoba2025towards} & $(H-s(t)+1)^2$ & Binary Square Patches & Binary Square Patches  & $\pdx$ \\
% PSPC-Flex \citep{niedoba2025towards} & $H^2$ & Binarized Sensitivity Maps & Binarized Sensitivity Maps  & $\pdx$ \\
% Lukoianov \citep{lukoianovlocality} & $d$ & Binarized Wiener Filter & Per-Dimension & $\pdx$ \\
% Ours & $H^2$ & Learned & Learned  & $\pdx$ \\
% \hline
% \end{tabular}

% \begin{tabular}{llll}
% \hline
% \textbf{Denoiser} & $\mathcal{Q}$ & $\mathcal{R}$ & $\mathcal{V}$ \\
% \hline
% Optimal & Identity & Identity & $\{\pd\}$ \\
% LS \citep{kamb2025analytic} & Binary Square Patches & Per-Pixel  & $\{\pd\}^{W\cdot H}$ \\
% ELS \citep{kamb2025analytic} & Binary Square Patches & Per-Pixel & $\{p_{T(x,y,s(t)(\mathcal{D})}\}_{x,y \in [W]\times[H]}$ \\
% PSPC-Square \citep{niedoba2025towards} & Binary Square Patches & Binary Square Patches  & $\{\pd\}^{(H-s(t) + 1)^2}$ \\
% PSPC-Flex \citep{niedoba2025towards}  & Binarized Sensitivity Maps & Binarized Sensitivity Maps  & $\{\pd\}^{W\cdot H}$ \\
% Lukoianov \citep{lukoianovlocality} & Binarized Wiener Filter & Per-Dimension & $\{\pd\}^d$ \\
% Ours & Learned & Learned  & $\{p_{\mathcal{D} \cup \mathcal{D}'}\}^{W\cdot H}$ \\
% \hline
% \end{tabular}

\begin{tabular}{llll}
\toprule
\textbf{Denoiser} & $\mathcal{Q}$ & $\mathcal{R}$ & $\mathcal{V}$ \\
\midrule
% Optimal & \{\mathbb{} & Identity & $\{\pd\}$ \\
LS \citep{kamb2025analytic} & $\left(\mathbf{p}\left(x,y,s_t\right) \mid x,y \in \mathcal{I}_0 \right)$ & $\left(\mathbf{p}\left(x,y, 1\right) \mid x,y \in \mathcal{I}_0\right)$   & $\left(\nu_\ell = \pd\right)_{\ell=1}^{|\mathcal{I}_0|}$ \\
ELS \citep{kamb2025analytic} & $\left(\mathbf{p}\left(x,y,s_t\right) \mid x,y \in \mathcal{I}_0 \right)$   & $\left(\mathbf{p}\left(x,y,1 \right) \mid x,y \in \mathcal{I}_0\right)$  & $\left(p_{\mathcal{T}_{x,y,s_t}(\mathcal{D})} \mid x,y \in \mathcal{I}_0 \right)$ \\
PSPC-Square \citep{niedoba2025towards} & $\left(\mathbf{p}\left(x, y,s_t\right) \mid x,y \in \mathcal{I}_{s_t}\right)$   & $\left(\mathbf{p}\left(x, y,s_t\right) \mid x,y \in \mathcal{I}_{s_t}\right)$   & $\left(\nu_\ell = \pd\right)_{\ell=1}^{|\mathcal{I}_{s_t}|}$ \\
PSPC-Flex \citep{niedoba2025towards}  & $(\mathbf{1}(\hat{\mathbf{G}_t}(x,y) < \tau_t) \mid x,y \in \mathcal{I}_0)$  &  $(\mathbf{1}(\hat{\mathbf{G}_t}(x,y) < \tau_t) \mid x,y \in \mathcal{I}_0)$  & $\left(\nu_\ell = \pd\right)_{\ell=1}^{|\mathcal{I}_0|}$ \\
Lukoianov \citep{lukoianovlocality} & $(\mathbf{1}(\hat{\mathbf{W}}_t(\ell) > \tau ))_{\ell=1}^d$ & $\left(\mathbf{e}_\ell\right)_{\ell=1}^d$ & $\left(\nu_\ell = \pd\right)_{\ell=1}^d$ \\
Ours & Fine-tuned & Fine-tuned  & $\left(\nu_\ell = p_{\mathcal{D} \cup \mathcal{D}'}\right)_{\ell=1}^{|\mathcal{I}_0|}$ \\
\midrule
\multicolumn{4}{l}{$\mathbf{p}(x,y,s) \in\{0,1\}^d $ produces square mask centred at pixel $(x,y)$ of size $s$, $\quad\mathbf{e}_\ell \in \{0,1\}^d$ is the $\ell$-th one-hot vector} \\
\multicolumn{4}{l}{$\mathbf{1}(\cdot) \in \{0,1\}^d$ is an indicator function, $\quad \hat{\mathbf{W}}_t(\ell) \in \mathbb{R}_+^d$ is the scaled Wiener filter row for dimension $\ell$} \\
\multicolumn{4}{l}{$\mathbf{\hat{G}_t(x,y)} \in \mathbb{R}_+^d$ is the cumulative gradient sensitivity map for $t$ at pixel $(x,y)$ \citep{niedoba2025towards}, $\quad s_t \in \mathbb{N}$ are patch sizes for time $t$}\\
\multicolumn{4}{l}{$\mathcal{I}_s = \{ x,y \mid x \in \{ \lfloor \frac s2\rfloor, \ldots, W - \lfloor \frac s2\rfloor - 1\}, y \in \{ \lfloor \frac s2\rfloor, \ldots, H - \lfloor \frac s2\rfloor - 1\} \}$, $\quad \tau_t \in \mathbb{R}_+^d$ are thresholds for time $t$} \\
\multicolumn{4}{l}{$\mathcal{T}_{x,y,s_t}$ is a equivariant translation set (\cref{eq:els_transforms}), $\quad \mathcal{D}'$ is an augmented dataset, (\cref{sec:aug})}\\
\bottomrule
\end{tabular}

        }
    }
    \label{tab:prior}
\end{table}
\label{sec:fpmc}
% Prompted by the unsuitability of the optimal denoiser as a model for network denoiser generalization, we propose a unified framework for analytical denoisers which approximate network behaviour through a weighted aggregation of posterior means. We refer to this model class as Filtered Posterior Mean Collections (FPMCs). 

Motivated by the diversity of prior methods which leverage posterior mean denoisers, we propose a framework which encompasses these methods and clarifies the common factors among their approaches. In the following section, we define the class of models which we refer to as filtered posterior mean collections (FPMCs).

% The foundation of an FPMC is a family of denoiser functions $\hat{\mu}(\bz, t ; q, \nu): \mathbb{R}^d \times \mathbb{R}_{>0} \rightarrow \mathbb{R}^d$ based on \emph{filtered} posterior expectations. Denoiser function instances are parameterized by a query precision vector $q \in \mathbb{R}_{\geq0}^d$, and a discrete source probability measure $\nu$ on $\mathbb{R}^d$ with finite support $\text{supp}(\nu)$. The form of each estimator is given by

The foundation of FPMCs are a family of denoiser functions $\hat{\mu}(\bz, t ; q, \nu): \mathbb{R}^d \times \mathbb{R}_{>0} \rightarrow \mathbb{R}^d$ based on filtered posterior expectations. Functions in this family calculate posterior expectations of the form
\begin{equation}
    \hat{\mu}(\bz, t; q, \nu) = \sum_{\bx \in \text{supp}(\nu)} \tilde{p}_t(\bx \mid \bz ; q, \nu) \bx, \label{eq:estimator}
\end{equation}
where the weights of sum are based upon a Gaussian likelihood
\begin{equation}
    \tilde{p}_t(\bz \mid \bx ; q) \propto \exp\left( \frac{(\alpha(t)\bx - \bz)^\top \text{diag}(q) (\alpha(t)\bx - \bz)}{-2\sigma(t)^2} \right), \label{eq:mod_ll}
\end{equation}
which, through Bayes' Rule, induces posterior probabilities
\begin{equation}
    \tilde{p}_t(\bx \mid \bz ; q, \nu) = \frac{\tilde{p}_t(\bz \mid \bx; q)\nu(\bx)}{\sum_{\bx' \in \text{supp}(\nu)} \tilde{p_t}(\bz \mid \bx'; q) \nu(\bx')}. \label{eq:mod_posterior}
\end{equation}
FPMC denoiser functions are parameterized by a query vector $q\in \mathbb{R}^d$, and a discrete source probability measure $\nu$ on $\mathbb{R}^d$ with finite support $\text{supp}(\nu)$. Intuitively, the query vector \emph{filters} $\hat{\mu}$ by adjusting the per-dimension importance of differences between $\bz$ and $\bx$. A $q$ vector with localized structure attends to local regions in each $\bx$, upweighting images which are most similar to $\bz$ within that region. By contrast, setting $q = \mathbf{I_d}$ recovers $\tilde{p}_t(\bz \mid \bx; q) = \ptzx$, which uniformly weights differences between $\bx$ and $\bz$ across dimensions. The source measure $\nu$ controls the distribution which is filtered. Selecting $\nu=p_\mathcal{D}$, along with $q=\mathbf{I}_d$ results in $\hat{\mu} = \mathbb{E}_{p_\mathcal{D}}[\bx \mid \bz, t]$.

In practice, FPMCs typically employ a collection of $L$ filtered denoisers, each specializing to specific spatial regions of $\bz$. We define the contribution of each specialist function to the final output through a response vector $r \in \mathbb{R}_{\geq0}^d$ which weights the dimensions of $\hat{\mu}$. A filtered posterior mean collection of $L$ denoiser functions is characterized by the choice of $q,r,$ and $\nu$ for each of its component estimators. Summarizing these paramaters into sequences $\mathcal{Q}=(q_\ell)_{\ell=1}^L$, $\mathcal{V} = (\nu_\ell)_{\ell=1}^L$, and $\mathcal{R}=(r_\ell)_{\ell=1}^L$, the output of an FPMC denoiser is given by 
\begin{equation}
    D_{\text{FPMC}}(\bz, t ; \mathcal{Q}, \mathcal{V}, \mathcal{R}) = \left(\sum_{\ell=1}^L r_\ell \odot \hat{\mu}(\bz, t ; q_\ell, \nu_\ell)\right) \oslash \left(\sum_{\ell=1}^L r_{\ell}\right), \label{eq:fpmc_denoiser}
\end{equation}
where $\odot$ and $\oslash$ denote element-wise multiplication and division respectively. Prior FPMC methods \citep{niedoba2025towards,kamb2025analytic,lukoianovlocality} are defined by their selections of $\Qset$,$\Vset$, and $\Rset$, which we summarize in \cref{tab:prior}.
\section{Improving FPMCs}
    \begin{figure}[t!]
    \centering
    \includegraphics[width=\linewidth]{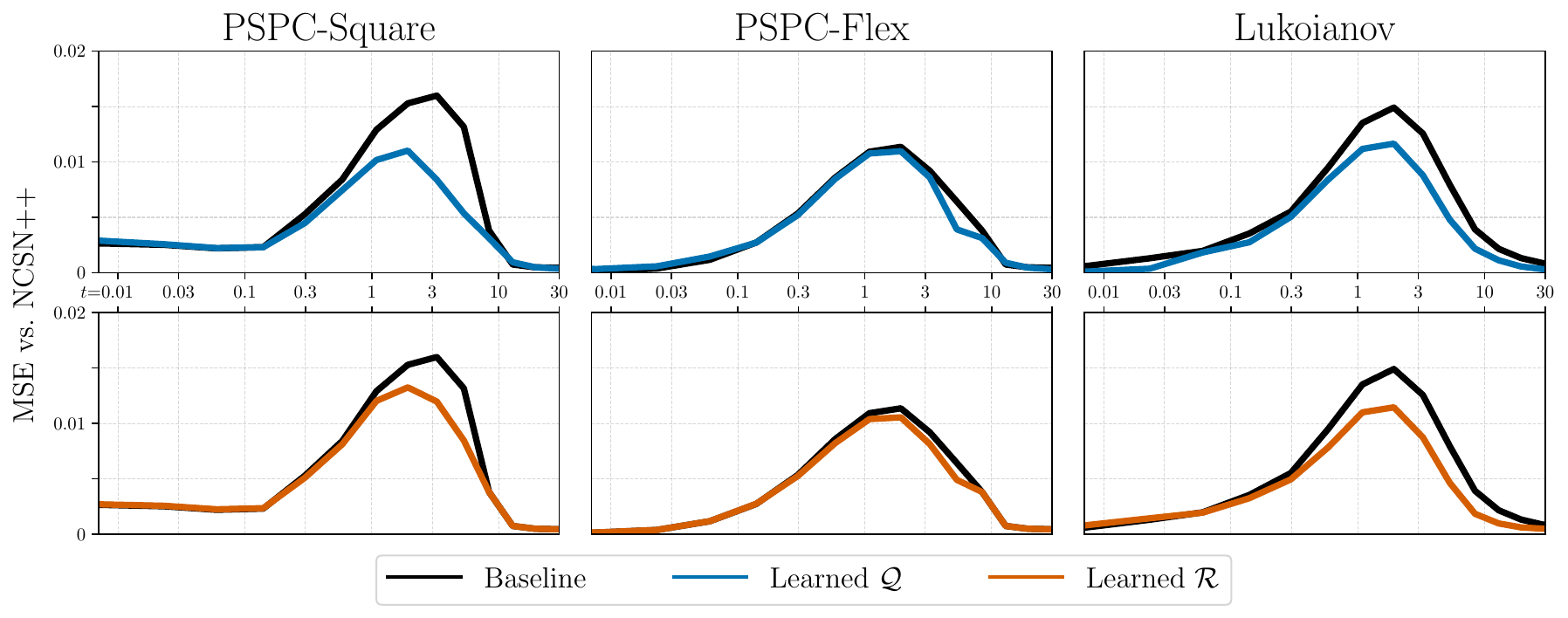}
    \caption{Effect of fine-tuning the $\Qset$ and $\Rset$ on denoiser error vs a CIFAR-10 DDPM++ denoiser for three prior FPMC methodologies. Mean squared error is reported over 1000 $\bz$ per $t$ value. \textbf{Top row:} Learning soft $\Qset$ (\textcolor[HTML]{0072b2}{\textbf{blue}}) consistently reduces the MSE of each method when compared against the corresponding binary $\Qset$ FPMC baseline (\textbf{black}). \textbf{Bottom row:} Learning soft $\Rset$ (\textcolor[HTML]{D55E00}{\textbf{orange}}) improves each FPMC versus the binary $\Rset$ baseline (\textbf{black}).}
    \label{fig:finetuning}
\end{figure}

\label{sec:improvements}
In this section, we investigate the principal design axes of FPMCs, identifying the limitations of prior choices and proposing improvements along each axis.
\subsection{Learning Soft $\Qset$}
\label{sec:softq}
The choice of $\Qset$ is the central decision of FPMC design. Prior art has primarily used $\Qset$ to imbue their FPMCs with a \emph{local} inductive bias -- supposing that model outputs for a given pixel should be correlated more strongly to spatially nearby input pixels. While motivations for this inductive bias are grounded in network gradients \citep{niedoba2025towards}, model architectures \citep{kamb2025analytic}, and covariance eigendecomposition \citep{lukoianovlocality}, all choices of $\Qset$ in prior methods share two key characteristics.

The first shared characteristic is a restriction of $q$ to $\{0,1\}^d$. This choice corresponds to $\hat{\mu}$ which are averaged over \emph{patches} of the dimensions of $\mathbb{R}^d$. As illustrated in \cref{fig:q_viz}, we argue that this restriction is inconsistent with the locality motivation of prior art as it requires every dimension within a patch to provide an equally weighted contribution to the posterior weights, regardless of spatial proximity.

Secondly, prior art uses manually tuned heuristics to derive the patches in $\Qset$. Both \citet{niedoba2025towards} and \citet{kamb2025analytic} draw inspiration from convolutional receptive fields, selecting $q$ corresponding to square image patches. However, hyperparameter tuning is needed to select the patch size $s_t$. Even when $\Qset$ is derived from data as in PSPC-Flex \citep{niedoba2025towards} and \citet{lukoianovlocality}, binary $q$ vectors require a clipping function and corresponding threshold hyperparameter $\tau_t$. 

We address the limitations of both characteristics. As presented in \cref{sec:fpmc}, we relax the space of $q$ from $\{0,1\}^d$ to $\mathbb{R}_{\geq0}^d$. Furthermore, we hypothesize that directly optimizing $\Qset$ against network denoiser functions
$D_\theta$ will lead to better FPMCs than selecting $\Qset$ through intuition and heuristics. We propose the straightforward optimization objective which mirrors the denoising score matching objective of \cref{eq:dsm}
\begin{equation}
    \mathcal{L}\left(\Qset, \Vset, \Rset\right) = \mathop{\mathbb{E}}_{\substack{t \sim p(t) \\\bx, \bz \sim \sim \ptzx p_\mathcal{D}(\bx), }}\left[\lambda(t)\left\lVert D_\theta\left(\bz, t\right) - D_{FPMC}\left(\bz, t ; \Qset, \Vset, \Rset\right)\right\lVert_2^2\right].\label{eq:obj}
\end{equation}
To minimize \cref{eq:obj} with respect to $\Qset$ the other FPMC design parameters must be selected. Specifically, we must choose response weights $\Rset$ and source distributions $\Vset$. To test the efficacy of optimizing $\Qset$ across a variety of $\Rset,$ and $\Vset$ choices, we apply our optimization procedure to three prior FPMCs  on the CIFAR-10 dataset \citep{krizhevsky2009learning}. In each case, we adopt $\Rset$ from the base method as specified in \cref{tab:prior}, and learn independent $\Qset$ per-$t$ which we initialize from the binary masks detailed in \cref{tab:prior}. The only other departure we make from prior methodology is to modify $\Vset$ during training per-batch such that $\nu(\bx)=0$ for the source training batch. We make this change to encourage generalization and avoid the trivial solution in which arbitrarily large precision vectors will collapse $\hat{\mu}$ to the source training batch $\bx$ in the small $t$ regime.

The top row of \cref{fig:finetuning} presents the effect of fine-tuning soft $\Qset$. For each of the three baseline FPMCs, we fine-tune $\Qset$ against a DDPM++ EDM denoiser \citep{karras2022elucidating}, and report the MSE of the fine-tuned and baseline FPMCs across varying $t$ values. For each $\sigma(t)$ we plot the MSE over 1000 $\bz$ points, sampled via $\bx \sim \mathcal{D}_\text{test}$ and $\bz \sim \ptzx$. From \cref{fig:finetuning}, learning a soft $\Qset$ consistently reduces FPMC error for all three methods, and is most helpful for intermediate $t$ values where FPMC error is highest. Among methods, fine-tuning $\Qset$ is most beneficial for PSPC-Square. We posit that this is because the data-driven patches of PSPC-Flex and \citet{lukoianovlocality} are closer to optimality than PSPC-Square, leaving less room for improvement. Taken together, the performance of fine-tuned FPMCs supports our claim that soft, optimized $\Qset$ are preferable to binary, heuristically derived alternatives.

\subsection{Learning Soft $\Rset$}
\label{sec:softr}

% Introduce the purpose of $\Rest$
The purpose of $\Rset$ is to define how the $L$ posterior mean estimators of an FPMC are aggregated to form a single output. As with the query precisions, all prior methods constrain their response weights to $r_\ell \in \{0,1\}^d$. Considering the methodologies of \citet{kamb2025analytic} and \citet{lukoianovlocality}, this is a natural restriction as both works propose $\Rset$ such that $\sum_\ell r_l = \mathbf{1}_d$. That is, each $\hat{\mu}_\ell$ estimator directly estimates a disjoint subset of the output dimensions. By contrast, \citet{niedoba2025towards} set $\Rset = \Qset$, and propose a simple uniform average over all overlapping output means. 

We suggest that neither approach is ideal. Partitioning $\mathbb{R}^d$ through $\Rset$ ignores that estimators with similar query precisions $q_\ell$ induce similar likelihoods $\tilde{p}(\bz \mid \bx; q_\ell)$, and therefore correlated $\hat{\mu}_\ell$. Aggregation of such correlated estimators may improve the final estimate. However, uniform aggregation ignores the degree of correlation between estimators, treating all overlapping estimators as equally informative. 

We argue that better values of $\Rset$ lie between these choices. As with $\Qset$, we propose a relaxation of $r_\ell$ from $\{0,1\}^d$ to $\mathbb{R}^d_{\geq0}$, and a data-driven selection of $\Rset$ through minimization of \cref{eq:obj} with respect to $\Rset$. We consider the effect of fine-tuning $\Rset$ from the baseline values described in \cref{tab:prior} for the same FPMCs considered in \cref{sec:softq}. Mirroring our optimization of $\Qset$, we freeze the other FPMC parameters to their baseline $\Qset $ values and use the modified $\Vset$ discussed in \cref{sec:softq}

The effect of fine-tuning soft $\Rset$, is shown in the bottom row of \cref{fig:finetuning}. Across all three FPMCs, fine-tuning $\Rset$ reduces FPMC error vs the DDPM++ denoiser. Among FPMCs, the most pronounced improvement is observed for \citet{lukoianovlocality}, whose $\Rset$ is comprised of one-hot $r$ vectors which prevent any mixing between posterior mean estimates. The observed improvement of \citet{lukoianovlocality} supports our claim that averaging related estimators can lead to improved FPMC performance.

% Introduce the choices which have been made by prior works & introduce limitations

% Introduce our hypothesis - that training R is beneficial because similar Q should have similar estimates of nearby pixels

% Introduce training objective

% Refer to figure and discuss.

\subsection{$\Vset$ Augmentations}
\label{sec:aug}

The choice of source image distributions $\Vset$ is the least explored axis of FPMC design. Among prior methods, only the ELS model \citep{kamb2025analytic} does not exclusively use $\nu=p_\mathcal{D}(\bx)$. Motivated by the equivariant inductive bias of convolutions, ELS differs from the related LS model solely through the use of per-estimator translation-augmented $\nu$. However, diffusion generalization consistency is not limited to purely convolutional architectures. Similar outputs are also produced by attention-enabled U-Nets and DiTs \citep{niedoba2025towards}, neither of which are translation equivariant due to their use of attention layers \citep{vaswani2017attention}.

Nevertheless, translational transformations are commonly employed in generative model training as a part of a broader class of data augmentation strategies. Such augmentations improve model training by expanding the empirical training distribution $p_\mathcal{D}(\bx)$ with plausible synthetic images to better approximate the true data generating distribution $p(\bx)$. We hypothesize that applying similar data augmentation to the source distributions of FPMC will also improve model performance. We therefore investigate augmenting $\Vset$ with other classical augmentation strategies, testing augmentation via translation, scaling, rotations, and reflections across the horizontal and vertical axes. Beyond classical augmentations, we additionally evaluate the impact of augmentation with additional diffusion model samples. Unlike classical augmentation strategies, diffusion samples represent the closest practical approximation to drawing additional i.i.d. samples from $p(\bx)$.

For our investigation, we generate augmented CIFAR-10 \cite{krizhevsky2009learning} source distributions through the union $\mathcal{D} \cup \mathcal{D}'$, where $\mathcal{D}'$ is comprised of images produced by a single augmentation strategy. We vary $|\mathcal{D}'|$ from $20\%$ to $2000\%$ of $|\mathcal{D}|$, corresponding to between 10,000 and 1,000,000 additional images. For each $|\mathcal{D}'|$ and corresponding $\nu=p_{\mathcal{D} \cup \mathcal{D}'}$, we evaluate a PSPC-Flex denoiser using 1000 test set images, and $\bz \sim \ptzx$ at three $\sigma(t)$. For each noise level, we report the relative change in FPMC approximation error versus DDPM++, expressed as a percentage of the unaugmented FPMC error.

\begin{figure}[t!]
    \centering
    \includegraphics[width=\linewidth]{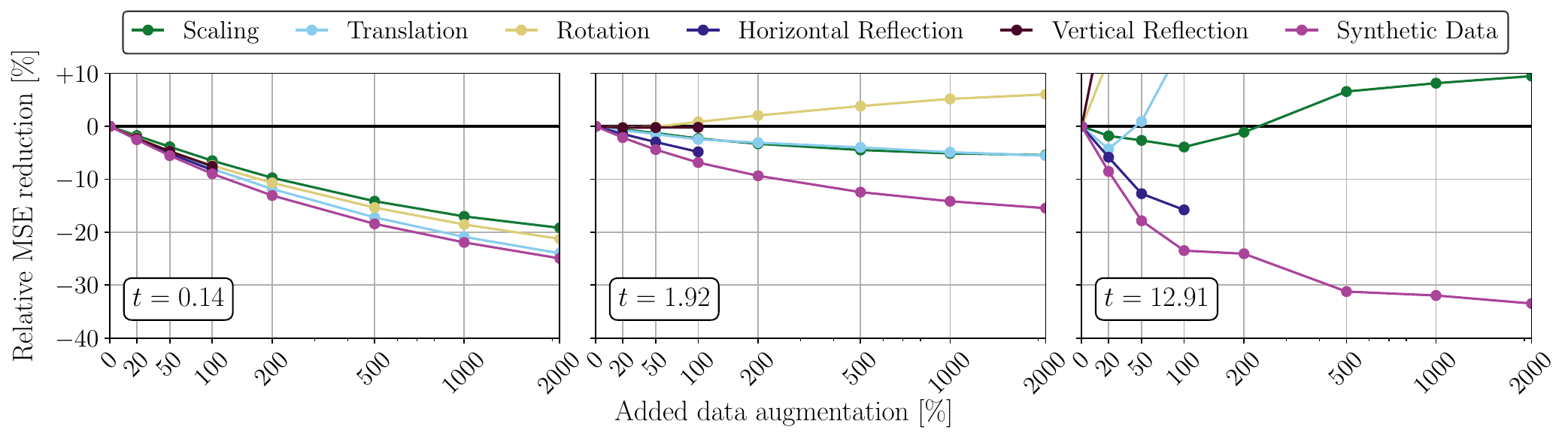}
    \caption{Comparison of the relative change in FPMC denoiser MSE vs a DDPM++ denoiser on the validation set of CIFAR-10 for three values of $t$. Investigating the impact of varying $\Vset$ by augmenting $\mathcal{D} $ with different methods and amounts of data augmentation, we find that only horizontal reflection and synthetic strategies consistently improve denoiser performance.}
    \label{fig:augmentations}
\end{figure}

\Cref{fig:augmentations} presents the effect of each augmentation strategy at each noise level. For low $t$, every augmentation method improves FPMC performance. However, as $t$ increases, some strategies become less effective, and even detrimental to FPMC performance. We find only horizontal reflection and synthetic augmentations are consistently effective at reducing FPMC error. We speculate that this is related to the growth of PSPC-Flex $\Qset$ patch size with $t$ \citep{niedoba2025towards}. At low $t$, the denoising task collapses to estimation of individual training images \citep{biroli2024dynamical}. At this scale, small patches may capture plausible local texture, even if the global structure is inconsistent with the target image. However, as patch sizes increase with $t$, globally implausible augmentations become harmful. Horizontal reflections and synthetic samples remain plausible at all scales, explaining their consistent benefit.

% Todo: Explain \cref{fig:augmentations}. High level story - at low noise level, all augmentations work pretty well. This is probably because at low noise level, the individual component denoisers of the FPMC are relatively local. At t increases, the means are capturing more global structure, and many transforms do not match the structure of the training distribution. At a high noise level, the network approximates the global mean of the dataset, which is corrupted by data augmentations. Xflip and synthetic data are the only transforms that don't mess with the data statistics too badly.
\section{Results}
    Building on the investigations of \cref{sec:improvements}, we propose a combined, general methodology for improving existing FPMCs through joint modifications to $\Qset$, $\Rset$, and $\Vset$. Following \cref{sec:softq,sec:softr}, we first relax the binary query precisions and response weights of the baseline FPMC to real-valued $q \in \mathbb{R}_{\geq0}^d$ and $r \in \mathbb{R}_{\geq0}^d$. Since the optimal choices of $\Qset$ and $\Rset$ are interdependent, we optimize both jointly against \cref{eq:obj}, producing a separate $\Qset$ and $\Rset$ for each $t$ value in the EDM sampling schedule \citep{karras2022elucidating}. In addition, we apply our conclusions from \cref{sec:aug} and expand $p_\mathcal{D}(\bx)$ with augmented datasets $\mathcal{D'}$ produced through horizontal reflection and synthetic augmentation strategies. For fair comparison, we fix the augmentation rate for each method at $100\%$ of training dataset size, the maximum augmentation achievable through horizontal reflection.

We apply our methodology to the PSPC-Flex denoiser \citep{niedoba2025towards}, selecting it for its strong performance in \cref{fig:finetuning}. However, we note that our methodology is generally applicable to any FPMC under the combined framework of \cref{sec:fpmc}. We evaluate our approach on three natural image datasets: CIFAR-10 \citep{krizhevsky2009learning}, FFHQ 64$\times$64 \cite{karras2019style}, and AFHQv2 64$\times$64 \cite{choi2020starganv2}. For each dataset, we jointly fine-tune the $\Qset$ and $\Rset$ of PSPC-Flex against publicly available DDPM++ EDM checkpoints \citep{karras2022elucidating}.

We compare the quality of samples produced by various denoisers via PF-ODE integration starting from a shared set of 1000 initial $\bz \sim \pi(\bz)$ against those produced by a NCSN++ EDM network \citep{karras2022elucidating}. We note that this network differs from the DDPM++ network used for PSPC-Flex finetuning and synthetic sample generation.

\begin{figure}[t!]
    \centering
    \includegraphics[width=\linewidth]{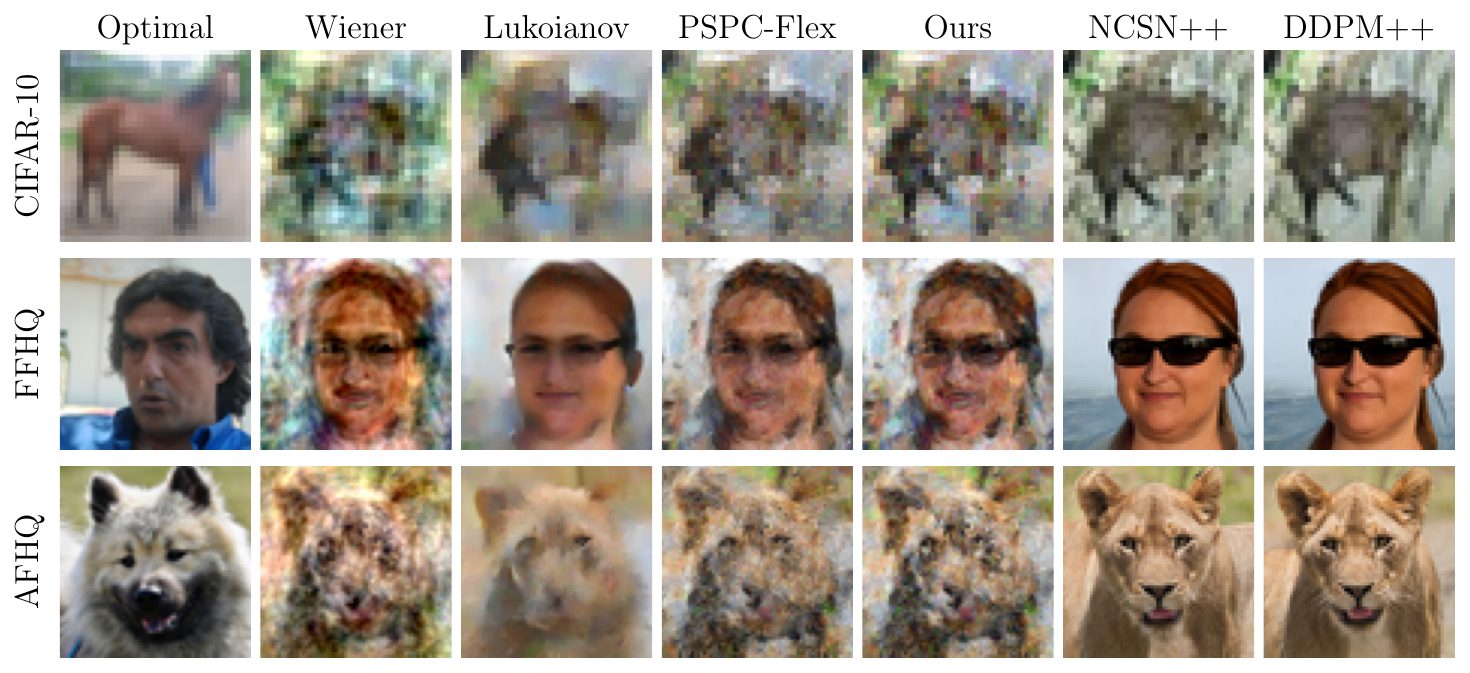}
    \caption{
         Comparison of samples generated by various denoisers with shared initial $\bz$ values. \textbf{Column 1:} The optimal denoiser memorizes training-set images, and does not match network samples. \textbf{Column 2:} The Wiener filter samples are better, but are over saturated. \textbf{Columns 3--5:} FPMC samples are most similar to network samples, capturing high-level structure but exhibiting high-frequency artifacts and lack of global coherence. Our FPMC ($\Qset/\Rset$ fine-tuned + synthetic) is qualitatively similar to its PSPC-Flex baseline. \textbf{Columns 6--7:} NCSN++ and DDPM++ diffusion models produce high quality, but remarkably similar samples.
    }
    \label{fig:samples}
\end{figure}

\Cref{tab:results}, compares the samples generated by each of the FPMC denoisers outlined in \cref{tab:prior}, as well as a Wiener filter, and DDPM++ network. For each FPMC, we re-implement the reported methodology under our shared framework. We report sample mean squared error (MSE) and $r^2$ vs NCSN++, along with standard errors. We do not evaluate ELS \citep{kamb2025analytic} on FFHQ 64$\times$64 or AFHQ $64\times64$ due to its computational complexity and poor results on face datasets in prior evaluation \citep{kamb2025analytic}. We provide qualitative comparisons for a subset of denoisers in \cref{fig:samples}, with additional samples in \cref{ap:additional}. From \cref{tab:results}, we find that applying our methodology to PSPC-Flex consistently improves sample similarity. In four of six metrics, we achieve state of the art performance, with near state of the art performance in the remaining two. Among augmentation strategies, synthetic augmentation outperforms horizontal reflection consistently. Examining \cref{fig:samples} however, we see that our FPMC is qualitatively very similar to its PSPC-Flex baseline, emphasizing that more research is required to close the gap to neural network samples.
\begin{table}[t!]
    \centering
    \caption{Quantitative comparison of samples drawn from identical initial conditions using varying denoisers to those produced by a NCSN++ diffusion model across three datasets. Best analytical denoiser within each section in \textbf{bold}, best overall analytical denoiser \underline{underlined}.}
    \resizebox{\textwidth}{!}{

\begin{tabular}{l cc cc cc}
\toprule
& \multicolumn{2}{c}{\textbf{CIFAR-10}} 
& \multicolumn{2}{c}{\textbf{FFHQ-64$\times$64}} 
& \multicolumn{2}{c}{\textbf{AFHQ-64$\times$64}} \\
\cmidrule(lr){2-3} \cmidrule(lr){4-5} \cmidrule(lr){6-7}
\textbf{Denoiser} 
& $r^2 \uparrow$ & MSE$\times10^2\downarrow$ 
& $r^2 \uparrow$ & MSE$\times10^2\downarrow$ 
& $r^2 \uparrow$ & MSE$\times10^2\downarrow$ \\
\midrule
\multicolumn{7}{l}{\small{\textit{Network Denoisers}}} \\
DDPM++ \citep{karras2022elucidating} & $0.93 \pm 0.00$ & $1.51 \pm 0.06$& $0.94 \pm 0.00$ & $1.19 \pm 0.04$& $0.85 \pm 0.00$ & $2.53 \pm 0.06$ \\
\midrule
\multicolumn{7}{l}{\small{\textit{Classical Denoisers}}} \\
Optimal & $-0.14 \pm 0.02$ & $17.90 \pm 0.24$& $0.06 \pm 0.01$ & $20.73 \pm 0.20$& $-0.23 \pm 0.01$ & $22.06 \pm 0.22$ \\
Wiener & $\mathbf{0.72 \pm 0.00}$ & $\mathbf{5.45 \pm 0.09}$& $\mathbf{0.73 \pm 0.00}$ & $\mathbf{6.11 \pm 0.07}$& $\underline{\mathbf{0.68 \pm 0.00}}$ & $\mathbf{6.28 \pm 0.07}$ \\
\midrule
\multicolumn{7}{l}{\small{\textit{Prior FPMCs}}} \\
LS \citep{kamb2025analytic} & $0.58 \pm 0.01$ & $7.17 \pm 0.11$& $0.72 \pm 0.00$ & $6.16 \pm 0.06$& $0.60 \pm 0.00$ & $6.97 \pm 0.07$ \\
ELS \citep{kamb2025analytic} & $0.66 \pm 0.01$ & $6.10 \pm 0.09$& - & -& - & - \\
PSPC-Square \citep{niedoba2025towards} & $0.57 \pm 0.01$ & $5.57 \pm 0.09$& $0.66 \pm 0.00$ & $5.69 \pm 0.07$& $0.48 \pm 0.01$ & $6.61 \pm 0.08$ \\
PSPC-Flex \citep{niedoba2025towards} & $0.67 \pm 0.01$ & $\mathbf{4.18 \pm 0.08}$& $\mathbf{0.74 \pm 0.00}$ & $\mathbf{\underline{4.48 \pm 0.05}}$ & $0.58 \pm 0.00$ & $\mathbf{5.65 \pm 0.07}$ \\
Lukoianov \citep{lukoianovlocality} & $\mathbf{0.69 \pm 0.01}$ & $4.32 \pm 0.07$& $0.69 \pm 0.00$ & $4.80 \pm 0.07$ & $0.38 \pm 0.01$ & $6.91 \pm 0.13$ \\
\midrule
\multicolumn{7}{l}{\small{\textit{Our Methodology}}} \\
Baseline (PSPC-Flex) & $0.67 \pm 0.01$ & $4.18 \pm 0.08$& $0.74 \pm 0.00$ & $\underline{\mathbf{4.48 \pm 0.05}}$& $0.58 \pm 0.00$ & $5.65 \pm 0.07$ \\
\hspace{0.5em} + Soft $\mathcal{Q}$/$\mathcal{R}$ & $0.73 \pm 0.00$ & $3.95 \pm 0.07$& $0.75 \pm 0.00$ & $4.63 \pm 0.05$& $0.65 \pm 0.00$ & $5.35 \pm 0.06$ \\
\hspace{0.5em} + Soft $\mathcal{Q}$/$\mathcal{R}$ + x-flip & $0.74 \pm 0.00$ & $3.84 \pm 0.07$& $0.75 \pm 0.00$ & $4.65 \pm 0.05$& $0.66 \pm 0.00$ & $5.38 \pm 0.06$ \\
\hspace{0.5em} + Soft $\mathcal{Q}$/$\mathcal{R}$ + synthetic & $\underline{\mathbf{0.76 \pm 0.00}}$ & $\underline{\mathbf{3.70 \pm 0.06}}$& $\underline{\mathbf{0.75 \pm 0.00}}$ & $4.57 \pm 0.05$& $\mathbf{0.67 \pm 0.00}$ & $\underline{\mathbf{5.22 \pm 0.06}}$ \\
\bottomrule
\end{tabular}

    }
    \label{tab:results}
\end{table}

\section{Related Work}
    \label{sec:related}
\paragraph{Diffusion Model Generalization} The denoising objective of \cref{eq:dsm} admits the optimal denoiser as its closed-form minimizer \citep{karras2022elucidating}, but this denoiser is incapable of generalization \citep{gu2023memorization}. \citet{yoon2023diffusion} demonstrate generalization occurs only when networks fail to memorize, a phenomenon which is driven by model capacity and dataset size \citep{yoon2023diffusion,gu2023memorization}. \citet{yi2023generalization} demonstrate that generalization emerges from biased estimators of the optimal score, which they attribute to optimization bias. \citet{zhang2024emergence} demonstrate that this bias is remarkably consistent across a variety of hyperparameters. A line of research has sought to investigate sources of this bias, such as model architecture \citep{kamb2025analytic, kadkhodaiegeneralization}, training distribution \cite{niedoba2025towards, song2025selective}, or loss stochasticity \cite{vastola2025generalization, niedoba2024nearest, bertrand2025closed}.

\paragraph{Analytic Denoisers} Many classical methods aim to recover images corrupted by moderate amounts of Gaussian noise \citep{roth2005fields, dabov2007image,buades2011non}. Of these, the Wiener filter is particularly relevant in the diffusion context. Research has found that these optimal linear denoisers are effective approximators of diffusion models \citep{wang2024unreasonable,li2024understanding}, echoing findings that show diffusion generalization can be approximated using shrinkage operations on the eigenspace of network jacobians instead of data covariance matrices \citep{kadkhodaiegeneralization}. \citet{lukoianovlocality} bridges these findings, connecting average network denoiser gradients and Wiener filters.

A distinct line of work proposes analytical denoisers which directly approximate diffusion model generalization through modifications to the optimal denoiser. One approach smooths the optimal score function to produce a generalizing denoiser \citep{scarvelis2023closed}. The focus of this work are methods based on the aggregation of multiple modified posterior means \citep{niedoba2025towards,kamb2025analytic,lukoianovlocality}. Our work, unifies these denoisers and elucidates their central design axes.

\section{Conclusions and Future Work}
    % High level conclusions

% \begin{itemize}
%         \item We introduce FPMCS, defined by $\Qset$, $\Rset$, and $\Vset$
%     \item We identify areas of improvement over prior work - specifically through relaxation of binary assumptions and augmentations of $\Vset$
%     \item When jointly applied to a PSPC-Flex baseline, our method generally improves FPMC sample similarity
%     \item While beneficial, FPMC performance still generally lags the quality of acutal diffusion models
% \end{itemize}

% Future Work
% \begin{itemize}
%     \item Full rank $q$ precision matrices
%     \item Non linear likelihoods
%     \item Adaptive likelihoods based on $\bz$
%     \item Learned source distributions
% \end{itemize}
\label{sec:conclusions}
The consistent generalization of neural-network image denoiser functions has motivated a growing line of research which aims to analytically approximate network denoisers. Among these, methods which aggregate posterior means over patches of the data have been particularly successful. In this work, we propose a unified filtered posterior mean collection framework which encompasses these fully network-free diffusion approximators. We investigate the design space of the model class, relaxing prior binary choices of $\Qset$ and $\Rset$ through optimization, and identifying beneficial strategies to augment $\Vset$. When applied jointly, these changes consistently improve FPMC sample similarity across three datasets. More broadly, we argue the systematic strategy employed by this work demonstrates the value of a unified FPMC framework which clearly delineates directions for future research.

Despite the improvements proposed in this work, there remain substantial limitations. Chief among these, there remains a substantial gap between FPMCs and network denoisers, at the level of both denoiser and sample similarity as demonstrated by \cref{fig:finetuning} and \cref{tab:results} respectively. Additionally, FPMCs are substantially less computationally efficient than their network counterparts. Regarding our contributions, although our method improves upon a PSPC-Flex baseline, in most cases, it does not improve MSE on FFHQ-$64\times64$, and falls short of state of the art for AFHQ $r^2$. Moreover, evaluation in this work is performed solely on datasets which generally have a single subject. More complex image content may require further advancements beyond what we have proposed.

% Maybe include optimization of V
The limitations discussed motivate several promising directions for FPMC development. The number of estimators $L$ in the FPMC is an implicit, unexplored design axis. Prior methods make widely varying choices of $L$, suggesting that systematic investigation of the optimal collection size may be warranted. In addition, relaxing the structure of the FPMC framework may produce further benefit. From \cref{eq:mod_ll}, current FPMCs have only investigated diagonal query precision matrices. More expressive precision matrices which incorporate important correlations between dimensions may lead to greater improvements. Going further, moving from Gaussian likelihoods to non-linear kernels may better approximate the non-linear behaviour of neural-network parameterized denoisers. Finally, while this work adapts the query precisions and response weights based on $t$, adapting FPMCs based on the input $\bz$ represents a natural extension which is likely to be especially useful for images which have discrete latent structure.

\section*{Acknowledgements}
We acknowledge the support of the Natural Sciences and Engineering Research Council of Canada (NSERC), the Alberta Machine Intelligence Institute (Amii) through the Canada CIFAR AI Chairs Program, Inverted AI, Mitacs, and Google. This research was enabled in part by technical support and computational resources provided by the Digital Research Alliance of Canada (alliancecan.ca), the Advanced Research Computing at the University of British Columbia (arc.ubc.ca), and Amazon Web Services.

\bibliographystyle{abbrvnat}
\bibliography{refs}

\newpage
\appendix
% \onecolumn
\section{Prior FPMCs}
    \label{ap:prior_methods}
In this section we restate the prior methodologies of \citet{niedoba2025towards}, \citet{kamb2025analytic}, and \citet{lukoianovlocality} under our combined FPMC framework.

\subsection{Patch Set Posterior Composites}

We will first derive a generic form for PSPC denoisers under our FPMC framework, before discussing specific the specific PSPC-Square and PSPC-Flex denoisers in \cref{ap:pspc_sq,ap:pspc_flex} respectively.

\citet{niedoba2025towards} express their patch set posterior composite denoiser using posterior means over patch datasets. They express a patch using cropping matrices $\mathbf{C} \in \{0,1\}^{n\times d}$ where $n\leq d$ is the dimension of the cropped patch. For any patch $\mathbf{C}$, they define patch likelihoods with $\bz_\mathbf{C} = \mathbf{C}\mathbf{z}$ and $\bx_\mathbf{C} = \mathbf{C}\bx$ as 
\begin{equation}
    p_t(\bz_\mathbf{C} \mid \bx_\mathbf{C}) = \mathcal{N}\left(\bz_\mathbf{C} ; \mathbf{x}_\mathbf{C}, t^2\mathbf{I}_d\right).
\end{equation}
In their work, they also adopt the parameterization of \citet{karras2022elucidating} with $\alpha(t)=1$ and  $\sigma(t)=t$. We can express their patch likelihoods more generally as
\begin{equation}
        p_t(\bz_\mathbf{C} \mid \bx_\mathbf{C}) = \mathcal{N}\left(\bz_\mathbf{C} ; \alpha(t)\mathbf{x}_\mathbf{C}, \sigma(t)^2\mathbf{I}_d\right)
\end{equation}
The log likelihood of this distribution satisfies
\begin{align}
    % \log p_t(\bz_\mathbf{C} \mid \bx_\mathbf{C}) &= \frac{\lVert\alpha(t) \bx_\mathbf{C} - \bz_\mathbf{C}\lVert_2^2}{-2\sigma(t)^2}  -\frac{n}{2} \log(2\pi\sigma(t)^2) \\
    % p_t(\bz_\mathbf{C} \mid \bx_\mathbf{C}) &= \exp\left(\frac{\lVert\alpha(t) \bx_\mathbf{C} - \bz_\mathbf{C}\lVert_2^2}{-2\sigma(t)^2}\right) \cdot\exp\left(\frac{n}{2} \log(2\pi\sigma(t)^2)\right)^{-1} \\
    p_t(\bz_\mathbf{C} \mid \bx_\mathbf{C}) &\propto \exp\left(\frac{\lVert\alpha(t) \bx_\mathbf{C} - \bz_\mathbf{C}\lVert_2^2}{-2\sigma(t)^2}\right)\\
    p_t(\bz_\mathbf{C} \mid \bx_\mathbf{C}) &\propto \exp\left(\frac{\lVert\alpha(t)\mathbf{C} \bx - \mathbf{C} \bz\lVert_2^2}{-2\sigma(t)^2}\right)\\
    p_t(\bz_\mathbf{C} \mid \bx_\mathbf{C}) &\propto \exp\left(\frac{\lVert\mathbf{C}\left(\alpha(t)\bx -\bz\right)\lVert_2^2}{-2\sigma(t)^2}\right) \\
    p_t(\bz_\mathbf{C} \mid \bx_\mathbf{C}) &\propto \exp\left(\frac{\left(\mathbf{C}\left(\alpha(t)\bx -\bz\right)\right)^\top\left(\mathbf{C}\left(\alpha(t)\bx -\bz\right)\right) }{-2\sigma(t)^2}\right) \\
    p_t(\bz_\mathbf{C} \mid \bx_\mathbf{C}) &\propto \exp\left(\frac{\left(\left(\alpha(t)\bx -\bz\right)\right)^\top \mathbf{C}^\top \mathbf{C}\left(\alpha(t)\bx -\bz\right) }{-2\sigma(t)^2}\right) \label{eq:CTC}
\end{align}
$\mathbf{C}^\top\mathbf{C}$ is a $d\times d$ matrix for which $(\mathbf{C}^\top\mathbf{C})_{i,j} = \mathbf{c}_i \cdot \mathbf{c_j}$ where $\mathbf{c}_i$ is the i-th column of $\mathbf{C}$. Since $\mathbf{C}$ is a cropping matrix its columns are orthonormal, which results in a diagonal $\mathbf{C}^T\mathbf{C}$ matrix. We can define $\mathbf{C}^T\mathbf{C} = \text{diag}(q_\mathbf{C})$ with $q_i = \mathbf{c}_i^T\mathbf{c}_i$ and $\mathbf{c}_i$ is the $i$-th column of $\mathbf{C}$. We have $q_\mathbf{C} \in \{0,1\}^d$, with non-zero entries for dimensions which are part of the cropped region. Together, the PSPC patch likelihood can be expressed as 
\begin{equation}
    p_t(\bz_\mathbf{C} \mid \bx_\mathbf{C}) \propto \exp\left(\frac{\left(\left(\alpha(t)\bx -\bz\right)\right)^\top \text{diag}\left(q_\mathbf{C}\right)\left(\alpha(t)\bx -\bz\right) }{-2\sigma(t)^2}\right),
\end{equation}
which is equivalent to the modified likelihood of our framework given by \cref{eq:mod_ll}, with $q=q_\mathbf{C}$.

Using their patch posterior likelihood, \citet{niedoba2025towards} define a patch posterior mean as 

\begin{equation}
    \mathop{\mathbb{E}}_{p_\mathcal{D}} \left[\bx_\mathbf{C} \mid \mathbf{z}_\mathbf{C}, t\right] = \sum_{\mathbf{x}^{(i)} \in \mathcal{D}} p_t(\bx_\mathbf{C}^{(i)} \mid \mathbf{z}_\mathbf{C}) \mathbf{x}_\mathbf{C}^{(i)}, \label{eq:patch_posterior_mean}
\end{equation}
with 
\begin{equation}
    p_t(\bx_\mathbf{C}^{(i)} \mid \mathbf{z}_\mathbf{C}) = \frac{p_t(\bz_\mathbf{C} \mid \bx^{(i)}_\mathbf{C})}{\sum_{\mathbf{x}^{(j)} \in \mathcal{D}}p_t(\bz_\mathbf{C} \mid \bx_\mathbf{C}^{(j)})}
\end{equation}
where $\bx^{(i)} \in \mathcal{D}$ is assumed. Substituting $p_t(\bz_\mathbf{C} \mid \bx^{(i)}_\mathbf{C}) = \tilde{p}_t(\bz \mid \bx^{(i)}; q_\mathbf{C})$, and defining $\nu(\bx) = p_\mathcal{D}(\bx) = \frac{1}{|\mathcal{D}|} \ \forall \bx \in \mathcal{D}$, with $\text{supp}(\nu) = \mathcal{D}$, we have

\begin{align}
    p_t(\bx_\mathbf{C}^{(i)} \mid \mathbf{z}_\mathbf{C}) &= \frac{\tilde{p}_t(\bz \mid \bx^{(i)})}{\sum_{\mathbf{x}^{(j)} \in \mathcal{D}}\tilde{p}_t(\bz \mid \bx^{(j)})} \\
    p_t(\bx_\mathbf{C}^{(i)} \mid \mathbf{z}_\mathbf{C}) &= \frac{\tilde{p}_t(\bz \mid \bx^{(i)}) \cdot \frac{1}{|\mathcal{D}|}}{\sum_{\mathbf{x}^{(j)} \in \mathcal{D}}\tilde{p}_t(\bz \mid \bx^{(j)})\cdot \frac{1}{|\mathcal{D}|}} \\
    p_t(\bx_\mathbf{C}^{(i)} \mid \mathbf{z}_\mathbf{C}) &= \frac{\tilde{p}_t(\bz \mid \bx^{(i)}) \nu(\bx^{(i)})}{\sum_{\mathbf{x}^{(j)} \in \mathcal{D}}\tilde{p}_t(\bz \mid \bx^{(j)})\nu(\bx^{(j)})} \\
    p_t(\bx_\mathbf{C}^{(i)} \mid \mathbf{z}_\mathbf{C}) &= \frac{\tilde{p}_t(\bz \mid \bx^{(i)}) \nu(\bx^{(i)})}{\sum_{\mathbf{x} \in \text{supp}(\nu)} \tilde{p}_t(\bz \mid \bx)\nu(\bx)}
\end{align}
Therefore, with choices $q=q_\mathbf{C}$ and $\nu = p_\mathcal{D}$, we recover the modified posterior of \cref{eq:mod_posterior}. Substituting this posterior into \cref{eq:patch_posterior_mean} yields
\begin{align}
    \mathop{\mathbb{E}}_{p_\mathcal{D}} \left[\bx_\mathbf{C} \mid \mathbf{z}_\mathbf{C}, t\right] &= \sum_{\mathbf{x}^{(i)} \in \mathcal{D}} \tilde{p}_t(\bx^{(i)} \mid \mathbf{z}; q_\mathbf{C}, p_\mathcal{D}) \mathbf{x}_\mathbf{C}^{(i)} \\
    \mathop{\mathbb{E}}_{p_\mathcal{D}} \left[\bx_\mathbf{C} \mid \mathbf{z}_\mathbf{C}, t\right] &= \sum_{\mathbf{x}^{(i)} \in \mathcal{D}} \tilde{p}_t(\bx^{(i)} \mid \mathbf{z}; q_\mathbf{C}, p_\mathcal{D}) \mathbf{C}\mathbf{x}^{(i)} \\
    \mathop{\mathbb{E}}_{p_\mathcal{D}} \left[\bx_\mathbf{C} \mid \mathbf{z}_\mathbf{C}, t\right] &= \mathbf{C}\sum_{\mathbf{x}^{(i)} \in \mathcal{D}} \tilde{p}_t(\bx^{(i)} \mid \mathbf{z}; q_\mathbf{C}, p_\mathcal{D}) \mathbf{x}^{(i)} \\
    \mathop{\mathbb{E}}_{p_\mathcal{D}} \left[\bx_\mathbf{C} \mid \mathbf{z}_\mathbf{C}, t\right] &= \mathbf{C}\sum_{\mathbf{x}^{(i)} \in \text{supp}(p_\mathcal{D})} \tilde{p}_t(\bx^{(i)} \mid \mathbf{z}; q_\mathbf{C}, p_\mathcal{D}) \mathbf{x}^{(i)} \\
    \mathop{\mathbb{E}}_{p_\mathcal{D}} \left[\bx_\mathbf{C} \mid \mathbf{z}_\mathbf{C}, t\right] &= \mathbf{C}\hat{\mu}(\bz, t; q_\mathbf{C}, p_\mathcal{D}). \label{eq:patch_posterior_mean_simplified}
\end{align}
Therefore, we can recover the patch posterior means of \citet{niedoba2025towards} using our modified posterior mean estimator defined in \cref{eq:estimator} with choices $q=q_\mathbf{C}$ and $\nu=p_\mathcal{D}$.

The final PSPC denoiser of \citet{niedoba2025towards} is an aggregation of a set of patch posterior means. they define their PSPC denoiser as 
\begin{equation}
        D\left(\bz, t, \mathcal{C} \right) = \bigg(\sum_{\mathbf{C} \in \mathcal{C}} \mathbf{C}^\top \mathbf{C} \bigg)^{-1}\sum_{\mathbf{C} \in \mathcal{C}} \mathbf{C}^\top \mathbb{E}\left[\bxi_\mathbf{C} \mid \bz_\mathbf{C}, t \right].\label{eq:patch_denoiser}
\end{equation}
As previously stated, for a cropping matrix with orthonormal rows, $\mathbf{C}^\top\mathbf{C} = \text{diag}(q_\mathbf{C})$. With this simplification, substituting \cref{eq:patch_posterior_mean_simplified} into \cref{eq:patch_denoiser} produces 
\begin{align}
        D\left(\bz, t, \mathcal{C} \right) &= \bigg(\sum_{\mathbf{C} \in \mathcal{C}} \mathbf{C}^\top \mathbf{C} \bigg)^{-1}\sum_{\mathbf{C} \in \mathcal{C}} \mathbf{C}^\top \mathbf{C} \hat{\mu}(\bz, t; q_\mathbf{C}, p_\mathcal{D}) \\
        D\left(\bz, t, \mathcal{C} \right) &= \sum_{\mathbf{C} \in \mathcal{C}} \text{diag}(q_\mathbf{C}) \hat{\mu}(\bz, t; q_\mathbf{C}, p_\mathcal{D}) \bigg(\sum_{\mathbf{C} \in \mathcal{C}} \text{diag}(q_\mathbf{C}) \bigg)^{-1}
\end{align}
Noting that multiplication by a diagonal matrix is equivalent to elementwise multiplication, and defining $r_\mathbf{C} = q_\mathbf{C}$, we have
\begin{equation}
    D\left(\bz, t, \mathcal{C} \right) = \sum_{\mathbf{C} \in \mathcal{C}} r_\mathbf{C} \odot \hat{\mu}(\bz, t; q_\mathbf{C}, p_\mathcal{D}) \oslash \bigg(\sum_{\mathbf{C} \in \mathcal{C}} r_\mathbf{C} \bigg),
\end{equation}
recovering the generic FPMC denoiser defined by \cref{eq:fpmc_denoiser} with $\Qset = (q_\mathbf{C} \mid \mathbf{C} \in \mathcal{C})$, $\Rset = (q_\mathbf{C} \mid \mathbf{C} \in \mathcal{C})$, and $\Vset = (\nu_\ell = p_\mathcal{D})_{\ell=1}^{|\mathcal{C}|}$. Next, we will discuss the two specific PSPC denoisers proposed by \citet{niedoba2025towards}, and define their $\Qset$ and $\Rset$.

\subsubsection{PSPC-Square}
\label{ap:pspc_sq}

PSPC-Square defines a square patch function $\mathbf{C}(x, y, s)$ which produces a cropping matrix corresponding to the square patch of an image with upper left corner at pixel $(x,y)$. They define the set of all overlapping patches with size $s$ as
\begin{equation}
    \mathcal{C}_s = \left\{\mathbf{C}(x,y,s) \mid x,y \in \{0, \ldots, W - s\} \times \{0, \ldots, H - s\}\right\}.\label{eq:pspc_cropping_set}
\end{equation} 
The size of patches in PSPC-Square is controlled by a patch schedule function $s(t): \mathbb{R}_{\geq0} \rightarrow \mathbb{N}$ which they manually tune through hyperparameter search. 

For simplicity, we define $q_\mathbf{C}$ directly using an indicator function. Let $\mathbf{p}(x,y,s): \{0,\ldots, W-1\} \times \{0, \ldots, H-1\} \times \mathbb{N} \rightarrow \{0,1\}^d$ be the indicator function which assigns a value of $1$ to pixels contained in the $s\times s$ square patch centred at $(x,y)$. That is $\forall \ i,j \in \{0, \ldots, W-1\} \times \{0, \ldots, H-1\}$ we have 
\begin{equation}
    \mathbf{p}(x,y,s)_{(i,j)} = \begin{cases}
        1 \quad \text{if} \ x  - \lfloor s/2 \rfloor \le i \leq x + \lfloor s/2 \rfloor  \wedge \ y - \lfloor s/2 \rfloor \leq j \leq y + \lfloor s/2 \rfloor \\
        0 \quad \text{otherwise},
    \end{cases} \label{eq:sq_patch_indicator}
\end{equation}
Where $\lfloor x \rfloor$ is the floor operator and $\wedge$ is the logical and operator. Additionally, let us define the set containing the centre $x,y$ coordinates of all valid $s\times s$ patches
\begin{equation}
    \mathcal{I}_s = \left\{ x,y \mid x \in \left\{ \lfloor \frac s2\rfloor, \ldots, W - \lfloor \frac s2\rfloor - 1\right\}, y \in \left\{ \lfloor \frac s2\rfloor, \ldots, H - \lfloor \frac s2\rfloor - 1\right\} \right\}.
\end{equation}

Using these definition, with the $\Qset$ for PSPC-Square becomes

\begin{equation}
    \Qset_{\text{PSPC-Square}} = \left(\mathbf{p}(x,y,s(t)) \mid x,y \in \mathcal{I}_{s(t)} \right)
\end{equation}

\subsubsection{PSPC-Flex}
\label{ap:pspc_flex}

PSPC-Flex applies a data-informed process to select their patch sets. Patch sets for a given $t$ are derived from channel averaged absolute gradients of a network denoiser at that noise level. For a given pixel $(x,y)$, \citet{niedoba2025towards} term these gradient sensitivity maps, and define them as 
\begin{equation}
    \mathbf{G}(x,y,t; D_\theta, p_\mathcal{D}) = \mathop{\mathbb{E}}_{\substack{\bx \sim p_\mathcal{D} \\ \bz \sim \ptzx}} \left[ \sum_{c=1}^3 \left| \nabla_\mathbf{z_c} D_\theta(\bz, t)_{x,y,c}\right|\right].
\end{equation}
We have augmented their original notation with a parameterization by a neural network denoiser and a data distribution for clarity. 

\citet{niedoba2025towards} convert the gradient sensitivities into cropping matrices through a cumulative thresholding function parameterized by a threshold value $\tau$. In their procedure, they construct cropping matrices $\mathbf{C}_\mathbf{G}$ implicitly, stating they select pixels in descending order until they satisfy the expression
\begin{equation}
    \sum \left(\mathbf{C}_G(x,y,t, \tau)\mathbf{G}(x,y,t ; D_\theta, p_\mathcal{D})\right) = \tau \sum \mathbf{G}(x,y,t ; D_\theta, p_\mathcal{D}) \label{eq:cumulative_thresh}
\end{equation}
Like PSPC-square, \citet{niedoba2025towards} derive a threshold schedule function $\tau(t) : \mathbb{R}_{\geq0} \rightarrow [0,1]$ which they tune through hyperparameter search.

Let $\texttt{cumsum}(\cdot)$ be a function which cumulatively sums the sorted elements of the input vector in descending order. Then, we define the cumulative gradient sensitivity map as
\begin{equation}
    \hat{\mathbf{G}}_t(x,y) = \texttt{cumsum}(\mathbf{G}(x,y,t))
\end{equation}

which we use to define the PSPC-Flex $\Qset$

\begin{equation}
    \Qset_{\text{PSPC-Flex}} = \left( \mathbf{1}\left(\hat{\mathbf{G}}_t\left(x,y\right) < \tau\left(t\right)\right) \mid x,y \in \mathcal{I}_0\right)
\end{equation}

Where $\mathbf{1}(\cdot) \in \{0,1\}^d$ is an indicator function.

\subsection{Local and Equivariant Local Score Machines}

We begin this section with a translation between our notation and the notation of \citet{kamb2025analytic}, given in \cref{tab:notation}

\begin{table}[t!]
    \centering
    \caption{General correspondence between our notation and \citet{kamb2025analytic}}
    \begin{tabular}{l c c}
    \hline
         \textbf{Concept} & \textbf{Kamb \& Ganguli} & \textbf{Ours} \\
    \hline
         Clean images &  $\varphi$ & $\bx$ \\
         Noised images & $\phi$ & $\bz$ \\
         Scale & $\sqrt{\bar{\alpha}_t}$ & $\alpha(t)$ \\
         Noise Scale & $(1 - \bar{\alpha}_t)$ & $\sigma(t)$ \\
         Conditional Probabilities & $\pi_t(\phi \mid \varphi)$ & $p_t(\bz \mid \bx)$ \\
         Marginal Probabilities & $\pi_t(\phi)$ & $p_t(\bz)$\\
         Posterior Probabilities & $W_t(\varphi \mid \phi)$ & $p_t(\bx \mid \bz)$ \\
         Prior & $\pi_T(\phi)$ & $\pi(\bz)$ \\
         Dataset & $\mathcal{D}$ & $\mathcal{D}$ \\
         Pixel Indexing & $\phi(x) \in \mathbb{R}^C$ & $\bx_{(x,y)} \in \mathbb{R}^C$ \\
         Local Region & $\Omega_x$ & $q \in \{0,1\}^d$ \\
         Local Region Restriction & $\phi_{\Omega_x} \in \mathbb{R}^{|\Omega_x|\times C}$ & $q \odot \bx$ \\
         Set of all patches & $P_{\Omega}(\mathcal{D}) $ & $-$ \\
         \hline
    \end{tabular}
    \label{tab:notation}
\end{table}

\citet{kamb2025analytic} define their local and equivariant score machine as estimators of the marginal score function. Through \cref{eq:tweedies}, recovery of the score function using a denoising function is possible through simple algebraic manipulation. For consistency with the FPMC framework, we will restate the LS and ELS score estimators in a denoiser function form.

\subsubsection{Local Score Machine}

The output of the LS denoiser under the notation of \cref{tab:notation} at pixel $(x,y)$ and a local region defined by a binary mask vector $q_{xy}$ which restricts $\bz$ and $\bx$ to a local region around $(x,y)$, is given by the equations
\begin{align}
    D_{\text{LS}}(\bz, t ; q_{xy})_{(x,y)} &= \sum_{\bx \in \mathcal{D}} p_t(q_{xy}\odot \bx \mid q_{xy} \odot \bz) \cdot \bx_{(x,y)} \label{eq:ls_denoiser}\\
    p_t(q_{xy}\odot \bx \mid q_{xy} \odot \bz) &= \frac{\mathcal{N}(q_{xy}\odot\bz ; \alpha(t) q_{xy}\odot\bx, \sigma(t)^2\mathbf{I})}{\sum_{\bx' \in \mathcal{D}} \mathcal{N}(q_{xy}\odot\bz ; \alpha(t) q_{xy}\odot\bx', \sigma(t)^2\mathbf{I})} \label{eq:ls_posterior}
\end{align}

For the Gaussian densities of \cref{eq:ls_posterior}, we have

\begin{align}
    \mathcal{N}(q_{xy}\odot\bz ; \alpha(t) q_{xy}\odot\bx, \sigma(t)^2\mathbf{I}) &\propto \exp\left(\frac{\lVert \alpha(t) q_{xy} \odot \bx - q_{xy}\odot \bz\lVert_2^2}{-2\sigma(t)^2}\right)\\
    \mathcal{N}(q_{xy}\odot\bz ; \alpha(t) q_{xy}\odot\bx, \sigma(t)^2\mathbf{I}) &\propto \exp\left(\frac{\lVert q_{xy}\odot \left(\alpha(t) \bx - \bz\right)\lVert_2^2}{-2\sigma(t)^2}\right) \\
    \mathcal{N}(q_{xy}\odot\bz ; \alpha(t) q_{xy}\odot\bx, \sigma(t)^2\mathbf{I}) &\propto \exp\left(\frac{\lVert \text{diag}\left(q_{xy}\right) \left(\alpha(t) \bx - \bz\right)\lVert_2^2}{-2\sigma(t)^2}\right) \\
    \mathcal{N}(q_{xy}\odot\bz ; \alpha(t) q_{xy}\odot\bx, \sigma(t)^2\mathbf{I}) &\propto \exp\left(\frac{\lVert \left(\text{diag}\left(q_{xy}\right) \left(\alpha(t) \bx - \bz\right)\right)^\top\left(\text{diag}\left(q_{xy}\right) \left(\alpha(t) \bx - \bz\right)\right)\lVert_2^2}{-2\sigma(t)^2}\right) \\
    \mathcal{N}(q_{xy}\odot\bz ; \alpha(t) q_{xy}\odot\bx, \sigma(t)^2\mathbf{I}) &\propto \exp\left(\frac{\lVert \left(\alpha(t) \bx - \bz\right)^\top \text{diag}\left(q_{xy}\right)^\top \text{diag} \left(q_{xy}\right) \left(\alpha(t) \bx - \bz\right)\lVert_2^2}{-2\sigma(t)^2}\right)
\end{align}

Since $q_{(x,y)}\in \{0,1\}^d$, and $\text{diag}\left(q_{(x,y)}\right)^\top = \text{diag}\left(q_{(x,y)}\right)$, we have $\text{diag}\left(q_{(x,y)}\right)^\top\text{diag}\left(q_{(x,y)}\right) = \text{diag}\left(q_{(x,y)}\right)$, and therefore $\mathcal{N}(q_{(x,y)}\odot\bz ; \alpha(t) q_{(x,y)}\odot\bx, \sigma(t)^2\mathbf{I}) = \tilde{p}_t(\bz \mid \bx ; q_{xy})$. Substituting into equation \cref{eq:ls_posterior} yields

\begin{align}
    p_t(q_{xy}\odot \bx \mid q_{xy} \odot \bz) &= \frac{\tilde{p}_t(\bz \mid \bx ; q_{xy})}{\sum_{\bx' \in \mathcal{D}} \tilde{p}_t(\bz \mid \bx' ; q_{xy})} \\
    &= \frac{\tilde{p}_t(\bz \mid \bx ; q_{xy})\cdot \frac{1}{|\mathcal{D}|}}{\sum_{\bx' \in \mathcal{D}} \tilde{p}_t(\bz \mid \bx' ; q_{xy}) \cdot\frac{1}{|\mathcal{D}|}} \\
    &= \frac{\tilde{p}_t(\bz \mid \bx ; q_{xy})\cdot p_\mathcal{D}(\bx)}{\sum_{\bx' \in \mathcal{D}} \tilde{p}_t(\bz \mid \bx' ; q_{xy}) p_\mathcal{D}(\bx')}\\
    &= \tilde{p}_t(\bx \mid \bz ; q_{xy}, p_\mathcal{D})
\end{align}

Thereby recovering the FPMC posterior defined in \cref{eq:mod_posterior} with selections $q=q_{xy}$, and $\nu=p_\mathcal{D}$. Substituting into \cref{eq:ls_denoiser} yields

\begin{equation}
    D_{\text{LS}}(\bz, t;q)_{(x,y)} = \sum_{\bx \in \mathcal{D}} \tilde{p}_t(\bx \mid \bz ; q_{xy}, p_\mathcal{D}) \cdot \bx_{(x,y)} \label{eq:ls_pixel_out_1}
\end{equation}

Using the patch indicator function defined in \cref{eq:sq_patch_indicator}, we can rewrite \cref{eq:ls_pixel_out_1} as
\begin{align}
     D_{\text{LS}}(\bz, t;q_{xy})_{(x,y)} &= \left(\sum_{\bx \in \mathcal{D}} \tilde{p}_t(\bx \mid \bz ; q_{xy}, p_\mathcal{D}) \cdot \left( \mathbf{p}(x,y, 1) \odot \bx \right)\right)_{(x,y)}  \\
     D_{\text{LS}}(\bz, t;q_{xy})_{(x,y)} &= \left(\mathbf{p}(x,y,1) \odot \left(\sum_{\bx \in \mathcal{D}} \tilde{p}_t(\bx \mid \bz ; q_{xy}, p_\mathcal{D}) \cdot \bx \right)\right)_{(x,y)} \\
     D_{\text{LS}}(\bz, t;q_{xy})_{(x,y)} &= \left(\mathbf{p}(x,y,1) \odot \hat{\mu}(\bz, t ; q_{xy}, p_\mathcal{D})\right)_{(x,y)},
\end{align}
demonstrating that the LS denoiser output for a pixel $(x,y)$ can be described using the modified FPMC posterior mean of \cref{eq:estimator}.

Next, we note that for $s=1$, we have $\sum_d \mathbf{p}(x,y, 1) = C \ \forall \ x,y \in \mathcal{I}_0$, where $C$ is the number of image channels. Additionally, we note that $\sum_{x,y \in \mathcal{I}_0} \mathbf{p}(x,y, 1) = \mathbf{1}$, indicating that each output dimension is estimated by a single FPMC posterior estimator. Therefore, we can drop the indexing operation, and express the general form of the LS denoiser for all pixels as

\begin{equation}
    D_{LS}(\bz, t) = \left(\sum_{x,y \in \mathcal{I}_0} \mathbf{p}(x,y,1) \odot \hat{\mu}(\bz, t; q_{xy}, p_\mathcal{D}) \right ) \oslash \left(\sum_{x,y \in \mathcal{I}_0}  \mathbf{p}(x,y,1)\right).
\end{equation}

Which recovers the general form of an FPMC denoiser given by \cref{eq:fpmc_denoiser}, with $\Qset = (q_{xy} \mid x,y \in \mathcal{I}_0)$, $\Rset = (\mathbf{p}(x,y,1) | x,y \in \mathcal{I}_0)$, and $\Vset = (\nu_\ell = p_\mathcal{D})_{\ell=1}^{|\mathcal{I}_0|}$.

Like PSPC-Flex, the LS denoiser is based upon square patches, with a patch size sechedule $s(t): \mathbb{R}_{\geq0} \rightarrow \mathbb{N}$ which is manually tuned through hyperparameter search. The only difference is that the LS machine utilizes zero padding for patches where the square patch exceeds the image boundary. As both $\bz$ and $\bx$ are zero-padded, these padding dimensions do not impact $\tilde{p}_t(\bz \mid \bx ; q_{xy})$. We can therefore equivalently represent $q_{xy}$ without using padding through the indicator defined by \cref{eq:sq_patch_indicator}, which will result in non-square patches for boundary overlaps. We have

\begin{equation}
    \Qset_{\text{LS}} = \left(\mathbf{p}(x,y,s(t)) \mid x,y \in \mathcal{I}_0 \right)
\end{equation}

\subsubsection{Equivariant Score Machine}

The equivariant local score machine has the same $\Qset$ and $\Rset$ as the local score machine
\begin{align}
    \Qset_{\text{ELS}} = \left(\mathbf{p}(x,y,s(t)) \mid x,y \in \mathcal{I}_0 \right) \\
    \Rset_{\text{ELS}} = \left(\mathbf{p}(x,y,1) \mid x,y \in \mathcal{I}_0 \right)
\end{align}

The only difference between the denoisers is the patches over which the posterior mean is computed. \citet{kamb2025analytic} use $p_\Omega(\mathcal{D})$ to refer to the set comprised of every $\Omega$ shaped region of every element of $\mathcal{D}$, and replace the sum over $p_\mathcal{D}$ in \cref{eq:ls_denoiser} with one over $p_\Omega(\mathcal{D})$. There are two ways to construct this set. For some patch size $s(t)$, one approach is to first produce the set of all possible square cropping matrices, and apply each matrix to every image of $\mathcal{D}$. Alternatively, the same set can be constructed by applying a single binary $q$ to the translated images produced through the cartesian product of all valid translations and $\mathcal{D}$. We define $\mathbf{T}_{ij}: \mathbb{R}^d \rightarrow \mathbb{R}^d$ to be the transformation which shifts an image $i$ pixels horizontally and $j$ pixels vertically, padding with zeros. 

Our next step is therefore to define the set of valid transforms for each $(x,y,s)$ tuple. In the boundary-broken ELS machine, there are four sets of valid translations which correspond to centre patches, horizontal edge patches, vertical edge patches and corner patches. We define $k=\lfloor \frac s2 \rfloor$ and $[A,B] =\{i\in \mathbb{Z} \mid A\leq i < B\} $ for convenience. Then, for $q = \mathbf{p}(x,y,s)$, the set of all valid translations in each set is
\begin{align}
    \mathcal{T}_{\text{centre}}(x,y,s) &= \left\{\mathbf{T}_{ij} \mid i,j \in [k - x, W - x - k] \times [k - y, H - y - k]\right\} \\
    \mathcal{T}_{\text{horizontal}}(x,y,s) &= \left\{\mathbf{T}_{i0} \mid i \in [k - x, W - x - k]\right\} \\
    \mathcal{T}_{\text{vertical}}(x,y,s) &= \left\{\mathbf{T}_{0j} \mid j \in [k - y, H - y - k]\right\} \\
    \mathcal{T}_{\text{corner}}(x,y,s) &= \left\{\mathbf{T}_{00}\right\}.
\end{align}
The correct translation set is determined by the patch centre $(x,y)$ and size $s$. Together, we define

\begin{equation}
    \mathcal{T}_{x,y,s} = \begin{cases}
        \mathcal{T}_{\text{centre}}(x,y,s) & \left(k \leq x \leq W - k + 1\right) \wedge  \left(k \leq y \leq H - k + 1\right), \\
        \mathcal{T}_{\text{horizontal}}(x,y,s) & \left(k \leq x \leq W - k + 1\right)  \wedge  \left(y < k\right), \\
        \mathcal{T}_{\text{horizontal}}(x,y,s) & \left(k \leq x \leq W - k + 1\right)  \wedge  \left(y >  H - k + 1\right), \\
        \mathcal{T}_{\text{vertical}}(x,y,s) & \left(x < k\right) \wedge  \left(k \leq y \leq H - k + 1\right), \\
        \mathcal{T}_{\text{vertical}}(x,y,s) & \left(x > W - k + 1\right) \wedge  \left(k \leq y \leq H - k + 1\right), \\
        \mathcal{T}_{\text{corner}}(x,y,s) & \text{otherwise}
    \end{cases} \label{eq:els_transforms}
\end{equation}

We define a position translation set augmented dataset as 
\begin{equation}
    \mathcal{T}(\mathcal{D}) = \left\{\mathbf{T}\bx \mid \mathbf{T}, \bx \in \mathcal{T} \times \mathcal{D} \right\},
\end{equation}
which we use to define the $\Vset$ for the boundary-broken equivariant local score denoiser

\begin{equation}
    \Vset_{\text{ELS}} = \left(p_{\mathcal{T}_{x,y,s(t)}(\mathcal{D})} \mid x,y \in \mathcal{I}_0 \right)
\end{equation}
where $s(t)$ is a tuned patch schedule hyperparameter matching $\Qset_{\text{ELS}}$.

\subsection{\citet{lukoianovlocality}}

\citet{lukoianovlocality} adopt the notational conventions $\bx = x_0$, $\bz = x_t$, with diffusion parameters $\alpha(t) = \sqrt\alpha_t$, and $\sigma(t) = \sqrt{1-\alpha_t}$ with $t \in [0,1]$. Their proposed denoiser function is given by

\begin{equation}
    \hat{f}(\bz, t) = \sum_{i=1}^{|\mathcal{D}|} \mathbf{w}_i(\bz, t) \odot \bx^{(i)}, \label{eq:luk_denoiser}
\end{equation}
where $\mathbf{w}(\bz, t): \mathbb{R}^d \times \mathbb{R}_{\geq0} \rightarrow \mathbb{R}^d_{\geq0}$ is a weighting function for dataset image $i$ whose $\ell$-th component is described by
\begin{equation}
    \mathbf{w}_i(\bz, t)_\ell = \frac{\exp \left(\frac{\alpha(t)^2}{-2\sigma(t)^2} \left\lVert \tilde{W}_t^\ell \odot \left(\frac{1}{\alpha(t)}\bz - \bx^{(i)}\right) \right\lVert_2^2 \right)}{\sum_{j=1}^{|\mathcal{D}|} \exp\left(\frac{\alpha(t)^2}{-2\sigma(t)^2} \left\lVert \tilde{W}_t^\ell \odot \left(\frac{1}{\alpha(t)}\bz - \bx^{(j)}\right) \right\lVert_2^2\right) } \label{eq:luk_restated}
\end{equation}

Where $\hat{W}^\ell_t \in \{0,1\}^d$ is a binary vector derived from the $\ell$-th row of the Wiener filter. Given the empirical covarince $\Sigma_\mathcal{D} = \mathbb{E}_\mathcal{D} \left[ \bx\bx^\top\right] - \mathbb{E}_\mathcal{D} \left[ \bx\right]\mathbb{E}_\mathcal{D}\left[ \bx\right]^\top$ of $\pdx$, which can be factorized as $\Sigma_\mathcal{D} =\mathbf{U} \text{diag}(\mathbf{\lambda}) \mathbf{U}^\top$ via singular-value decomposition the Wiener filter matrix for time $t$ is

\begin{equation}
    \mathbf{W}_t = \frac{1}{\alpha(t)} \mathbf{U} \text{diag}\left(\frac{\alpha(t)^2\lambda}{\alpha(t)^2\lambda+\sigma(t)^2}\right)\mathbf{U}^\top
\end{equation}

To obtain query precisions, \citet{lukoianovlocality} first scale the Wiener filter matrix through row-wise scaling. Defining $\mathbf{W}_t(\ell)$ as the $\ell$-th row of the Wiener Filter, they produce a rescaled matrix through
\begin{equation}
    \hat{\mathbf{W}}_t(\ell) = \frac{\mathbf{W}_t(\ell)}{\max \mathbf{W}_t(\ell)},
\end{equation}
which they convert into binary vectors using the threshold operation 
\begin{equation}
    \tilde{W}_t^\ell = \mathbf{1}\left(\hat{\mathbf{W}}_t(\ell) > \tau\right)
\end{equation}
Where $\mathbf{1}(\cdot): \mathbb{R}^d \rightarrow \{0,1\}^d$ is an indicator function, and $\tau$ is constant threshold which is manually tuned through hyperparameter search.

Examining the numerator of \cref{eq:luk_restated}, we have
\begin{align}
    &\exp \left( \frac{\alpha(t)^2}{-2\sigma(t)^2} \left\lVert \tilde{W}_t^\ell \odot \left(\frac{1}{\alpha(t)}\bz - \bx^{(i)}\right) \right\rVert_2^2 \right) \\
    &= \exp \left(\frac{\left\lVert \tilde{W}_t^\ell \odot \left(\bz - \alpha(t)\bx^{(i)}\right) \right\lVert_2^2}{-2\sigma(t)^2} \right) \\
    &=     \exp \left(\frac{\left( \tilde{W}_t^\ell \odot\left(\bz - \alpha(t)\bx^{(i)}\right) \right)^\top\left( \tilde{W}_t^\ell \odot \left(\bz - \alpha(t)\bx^{(i)}\right) \right)}{-2\sigma(t)^2} \right)\\
    &=     \exp \left(\frac{\left(\bz - \alpha(t)\bx^{(i)}\right)^\top \text{diag}\left(\tilde{W}_t^\ell\right) ^\top \text{diag}\left(\tilde{W}_t^\ell\right) \left(\bz - \alpha(t)\bx^{(i)}\right)}{-2\sigma(t)^2} \right) \\
    &=     \exp \left(\frac{\left(\bz - \alpha(t)\bx^{(i)}\right)^\top \text{diag}\left(\tilde{W}_t^\ell\right) \left(\bz - \alpha(t)\bx^{(i)}\right)}{-2\sigma(t)^2} \right)\\ 
    &\propto \tilde{p}_t(\bz \mid \bx^{(i)} ; \tilde{W}_t^\ell)  \label{eq:luk_likelihood}
\end{align}

Choosing $q_\ell = \tilde{W}_t^\ell$, \cref{eq:luk_likelihood} is now matches the FPMC filtered likelihood defined in \cref{eq:mod_ll}. Substituting into \cref{eq:luk_restated} yields

\begin{align}
    \mathbf{w}_i(\bz, t)_\ell &= \frac{\tilde{p}_t(\bz \mid\bx^{(i)}; \tilde{W}_t^\ell)}{\sum_{j=1}^{|\mathcal{D}|} \tilde{p}_t(\bz \mid\bx^{(j)} ; \hat{W}_t^\ell)} \\
    \mathbf{w}_i(\bz, t)_\ell &= \frac{\tilde{p}_t(\bz \mid\bx^{(i)}; \tilde{W}_t^\ell) \cdot{\frac{1}{|\mathcal{D}|}}}{\sum_{j=1}^{|\mathcal{D}|} \tilde{p}_t(\bz \mid\bx^{(j)} ; \tilde{W}_t^\ell) \cdot {\frac{1}{|\mathcal{D}|}}} \\
    \mathbf{w}_i(\bz, t)_\ell &= \frac{\tilde{p}_t(\bz \mid\bx^{(i)}; \tilde{W}_t^\ell) \pd(\bx^{(i)})}{\sum_{j=1}^{|\mathcal{D}|} \tilde{p}_t(\bz \mid\bx^{(j)} ; \tilde{W}_t^\ell) \pd(\bx^{(j)})} \\
    \mathbf{w}_i(\bz, t)_\ell &= \tilde{p}_t(\bx^{(i)} \mid \bz ; \tilde{W}_t^\ell, \pd)
\end{align}

Let $\mathbf{e}_\ell \in \{0,1\}^d$ be a one-hot vector on dimension $\ell$. We can then rewrite \cref{eq:luk_denoiser} as 

\begin{align}
    \hat{f}(\bz, t) &= \sum_{\ell=1}^d \mathbf{e}_\ell \odot \sum_{i=1}^{|\mathcal{D}|} \mathbf{w}_i(\bz, t) \odot \bx^{(i)} \\
    \hat{f}(\bz, t)  &= \sum_{\ell=1}^d  \mathbf{e}_\ell \odot \sum_{i=1}^{|\mathcal{D}|} \mathbf{w}_i(\bz, t)_\ell \odot \bx^{(i)} \\
    \hat{f}(\bz, t)  &= \sum_{\ell=1}^d  \mathbf{e}_\ell \odot \hat{\mu}(\bz, t; \hat{W}_t^\ell, \pd) \\
    \hat{f}(\bz, t)  &= \left( \sum_{\ell=1}^d  \mathbf{e}_\ell \odot \hat{\mu}(\bz, t; \hat{W}_t^\ell, \pd) \right) \oslash  \left( \sum_{i=1}^{|\mathcal{D}|} \right)\\
\end{align}
Therefore, $\hat{f}$ is an FPMC denoiser under the choices 
\begin{align}
    \Qset_{L} &= \left(\mathbf{1}\left(\hat{\mathbf{W}}_t(\ell) > \tau\right)\right)_{\ell=1}^d \\
    \Rset_L &= \left(\mathbf{e}_\ell \right)_{\ell=1}^d \\
    \Vset_L &= (\nu_\ell = \pd)_{\ell=1}^d.
\end{align}

\section{Augmentation Strategies}
    In this section, we detail the augmentation utilized in \cref{sec:aug}. We first give an overview of how we generate our agumented datasets, and then detail each agumentation strategy. For the classical augmentation strategies, we use the augmentation pipeline of \citet{karras2022elucidating}, which was adapted from \citet{karras2020training}. The augmentation pipeline is a stochastic generator which produces tuples of augmented images and labels which describe the parameters used to generate the augmentation transformation. Although the augmentation pipeline can be configured produce a composition of multiple geometric augmentations, we restrict our investigation to generating images with one augmentation strategy at a time.

For each augmentation strategy, excluding the reflection transforms where only one augmentation per source image is possible, we first generate 1 million augmented images by sampling 20 augmentations per original training set image. When sampling augmented images using classical augmentation strategies, we keep a hash of the generated augmentation labels and discard any duplicate augmentations. From the original augmented dataset of 1 million images, we produce smaller augmentation sets through random sub sampling. For dataset sizes which are integer multiples of $|\mathcal{D}|$, we randomly select an equal amount of augmented images per source training image. For augmentation sets which are smaller than the original dataset size, we select one augmented image each for a subset of the original training images. The following augmentation strategies were explored

\paragraph{Horizontal Reflection} For this reflection, we simply flip the image horizontally.
\paragraph{Vertical Reflection} For this reflection, we simply flip the image vertically.
\paragraph{Translation} We translate images according to random offsets sampled as
\begin{align}
    \delta_x &= \epsilon_x\cdot \frac{W}{8}, \quad \epsilon_x\sim\mathcal{N}(0,1) \\
    \delta_y &= \epsilon_y\cdot \frac{H}{8}, \quad \epsilon_y\sim\mathcal{N}(0,1).
\end{align}
Translated images are padded with reflection padding post-transformation.
\paragraph{Rotation} The amount of rotation expressed in radians is randomly sampled from the uniform distribution
\begin{equation}
    \theta \sim \mathcal{U}(-\pi, \pi)
\end{equation}
Like translations, images are padded with reflection padding to retain the original shape of the image post transformation.
\paragraph{Scaling} Scaling up or downsamples the image to resolution $sH \times sW$. The amount of scaling is sampled from the log normal distribution
\begin{equation}
    \ln(s) \sim \mathcal{N}(0,0.2)
\end{equation}
and reflection padding is used when $s < 1$.

\paragraph{Synthetic} Unlike prior augmentations, synthetic augmentations are not produced through transformations of the source dataset images. Instead, sample our synthetic dataset using the EDM DDPM++ denoiser, and Huen sampling, following the same sampling procedure outlined in \cref{ap:sampling}. Importantly, while we generate our synthetic samples from using PF-ODE integration, we are careful to use a different random seed. Therefore sampling is initialized from a different set of initial $\bz \sim \pi(\bz)$ as the seeds used sample evaluation in \cref{tab:results}.

\section{Experimental Details}
    \label{ap:exp_details}

\subsection{FPMC Hyperparameters}

Many of the FPMCs outlined in \cref{tab:prior} have specific schedules of hyperparameters. These include the patch size schedule $s(t)$ for LS, ELS \citep{kamb2025analytic} and PSPC-Square \citep{niedoba2025towards}, the threshold schedule $\tau(t)$ for PSPC-Flex \citep{niedoba2025towards} and the constant threshold $\tau$ used by \citet{lukoianovlocality}. 

For PSPC methods, as they utilize the same EDM network denoisers and diffusion process $\alpha(t)$ and $\sigma(t)$, we adopt the hyperparameter schedules reported in their original work. For \citet{lukoianovlocality}, we report their recommended threshold value for $\tau=0.02$, except for CIFAR-10, for which we use $\tau=0.05$, following the findings of appendix B.1 of their work. However, when writing this paper, we did evaluate $\tau=0.05$ for FFHQ and AFHQ. The performance of these hyperparameter choices are presented in \cref{tab:tau}
% \mathbf{0.78 \pm 0.00}}$ & $\underline{\mathbf{4.24 \pm 0.05}

\begin{table}[h!]
    \centering
    \caption{Comparison of \citet{lukoianovlocality} denoiser performance with reported hyperparameter $\tau=0.02$ and alternative hyperparameter $\tau=0.05$}
    \begin{tabular}{l cc cc}
    \toprule
    
    & \multicolumn{2}{c}{\textbf{FFHQ-64$\times$64}} 
    & \multicolumn{2}{c}{\textbf{AFHQ-64$\times$64}} \\
    \cmidrule(lr){2-3} \cmidrule(lr){4-5}
    \textbf{Denoiser} 
    & $r^2 \uparrow$ & MSE $\times10^2\downarrow$ 
    & $r^2 \uparrow$ & MSE $\times10^2\downarrow$ \\
    \midrule
    \citet{lukoianovlocality}, $\tau=0.02$ & $0.69 \pm 0.00$ & $4.80 \pm 0.07$ & $0.38 \pm 0.01$ & $6.91 \pm 0.13$ \\
    \citet{lukoianovlocality}, $\tau=0.05$ & $\mathbf{0.78 \pm 0.00}$ & $\mathbf{4.24 \pm 0.05}$ & $\mathbf{0.67 \pm 0.00}$ & $\mathbf{5.12 \pm 0.06}$ \\
    \bottomrule
    \end{tabular}
    \label{tab:tau}
\end{table}

For LS and ELS, we performed hyperparameter search to find the best $s(t)$, sweeping over odd patch sizes $\{3,5,7,9,11,15,19,23,27,31\}$ for CIFAR-10 and all odd patch sizes between 3 and 63 for AFHQ-$64\times64$ and FFHQ-$64\times64$. Due to the computational complexity of ELS, we only evaluated it on CIFAR-10 and did not perform a hyperparameter sweep for FFHQ or AFHQ.

We evaluated each patch size over a common set of 1000 $\bz \sim \ptzx \pdx$ for each $t$ value in the EDM sampling schedule for the relevant dataset. We sample $\bx$ from the test set of CIFAR-10, and a synthetic set of 1000 DDPM++ samples for AFHQ-$64\times64$ and FFHQ-$64\times64$. At each $t$, we selected the patch size which had the lowest error vs the EDM DDPM++ denoiser. 

A summary of the relevant hyperparameter schedules for the five FPMCs which we evaluated are given in \cref{tab:cifar_schedules,tab:ffhq_schedules,tab:afhq_schedules}.

\begin{table}[h!]
    \centering
    \caption{Hyperparameter schedules for various FPMCs for the CIFAR-10 dataset \citep{krizhevsky2009learning}, and EDM sampling schedule\citep{karras2022elucidating}}
        \begin{tabular}{l l | c c c c c}
        \hline
        \multirow{2}{*}{Step} & \multirow{2}{*}{$t$} & PSPC-Square & LS & ELS & PSPC-Flex & Lukoianov \\
        & & $s(t)$ & $s(t)$ & $s(t)$ & $\tau(t)$ & $\tau$ \\
        \hline
        0 & 80.0 & 32 & 31 & 31 & 1. & 0.05 \\
        1 & 57.6 & 32 & 31 & 31 & 1. & 0.05 \\
        2 & 40.8 & 32 & 31 & 31 & 1. & 0.05 \\
        3 & 28.4 & 32 & 31 & 31 & 1. & 0.05 \\
        4 & 19.4 & 32 & 31 & 31 & 1. & 0.05 \\
        5 & 12.9 & 32 & 31 & 31 & 1. & 0.05 \\
        6 & 8.40 & 32 & 29 & 31 & 1. & 0.05 \\
        7 & 5.32 & 23 & 25 & 27 & 0.7 & 0.05 \\
        8 & 3.26 & 15 & 15 & 19 & 0.7 & 0.05 \\
        9 & 1.92 & 11 & 11 & 15 & 0.5 & 0.05 \\
        10 & 1.09 & 7 & 7 & 9 & 0.4 & 0.05 \\
        11 & 0.585 & 5 & 5 & 7 & 0.4 & 0.05 \\
        12 & 0.296 & 3 & 3 & 5 & 0.4 & 0.05 \\
        13 & 0.140 & 3 & 3 & 3 & 0.4 & 0.05 \\
        14 & 0.060 & 3 & 3 & 3 & 0.3 & 0.05 \\
        15 & 0.023 & 3 & 3 & 3 & 0.3 & 0.05 \\
        16 & 0.008 & 3 & 3 & 3 & 0.3 & 0.05 \\
        17 & 0.002 & 3 & 3 & 3 & 0.3 & 0.05 \\
        \hline
    \end{tabular}
    \label{tab:cifar_schedules}
\end{table}

\begin{table}[h!]
    \centering
    \caption{Hyperparameter schedules for various FPMCs for the FFHQ-$64\times64$ dataset \citep{karras2019style}, and EDM sampling schedule\citep{karras2022elucidating}}
    \begin{tabular}{l l | c c c c}
    \hline
    \multirow{2}{*}{Step} & \multirow{2}{*}{$t$} & PSPC-Square & LS & PSPC-Flex & Lukoianov \\
    & & $s(t)$ & $s(t)$ & $\tau(t)$ & $\tau$ \\
    \hline
    0 & 80 & 64 & 63 & 1.00 & 0.02 \\
    1 & 69.5 & 64 & 63 & 1.00 & 0.02 \\
    2 & 60.1 & 64 & 63 & 1.00 & 0.02 \\
    3 & 51.9 & 64 & 63 & 1.00 & 0.02 \\
    4 & 44.6 & 64 & 63 & 1.00 & 0.02 \\
    5 & 38.3 & 64 & 63 & 1.00 & 0.02 \\
    6 & 32.7 & 64 & 63 & 1.00 & 0.02 \\
    7 & 27.8 & 64 & 63 & 1.00 & 0.02 \\
    8 & 23.6 & 64 & 63 & 1.00 & 0.02 \\
    9 & 19.9 & 64 & 63 & 1.00 & 0.02 \\
    10 & 16.8 & 64 & 63 & 1.00 & 0.02 \\
    11 & 14.1 & 64 & 55 & 0.80 & 0.02 \\
    12 & 11.7 & 43 & 51 & 0.70 & 0.02 \\
    13 & 9.72 & 43 & 45 & 0.60 & 0.02 \\
    14 & 8.03 & 43 & 39 & 0.60 & 0.02 \\
    15 & 6.59 & 35 & 35 & 0.60 & 0.02 \\
    16 & 5.38 & 35 & 31 & 0.60 & 0.02 \\
    17 & 4.37 & 27 & 29 & 0.55 & 0.02 \\
    18 & 3.52 & 23 & 25 & 0.50 & 0.02 \\
    19 & 2.82 & 19 & 21 & 0.45 & 0.02 \\
    20 & 2.24 & 15 & 17 & 0.40 & 0.02 \\
    21 & 1.77 & 15 & 15 & 0.40 & 0.02 \\
    22 & 1.38 & 11 & 11 & 0.40 & 0.02 \\
    23 & 1.07 & 9 & 7 & 0.35 & 0.02 \\
    24 & 0.823 & 9 & 7 & 0.30 & 0.02 \\
    25 & 0.625 & 7 & 7 & 0.30 & 0.02 \\
    26 & 0.470 & 5 & 7 & 0.30 & 0.02 \\
    27 & 0.349 & 5 & 5 & 0.30 & 0.02 \\
    28 & 0.256 & 3 & 3 & 0.30 & 0.02 \\
    29 & 0.185 & 3 & 3 & 0.30 & 0.02 \\
    30 & 0.131 & 3 & 3 & 0.30 & 0.02 \\
    31 & 0.092 & 3 & 3 & 0.30 & 0.02 \\
    32 & 0.063 & 3 & 3 & 0.30 & 0.02 \\
    33 & 0.042 & 3 & 3 & 0.30 & 0.02 \\
    34 & 0.028 & 3 & 3 & 0.30 & 0.02 \\
    35 & 0.018 & 3 & 3 & 0.30 & 0.02 \\
    36 & 0.011 & 3 & 3 & 0.30 & 0.02 \\
    37 & 0.006 & 3 & 3 & 0.30 & 0.02 \\
    38 & 0.004 & 3 & 3 & 0.30 & 0.02 \\
    39 & 0.002 & 3 & 3 & 0.30 & 0.02 \\
    \hline
\end{tabular}
    \label{tab:ffhq_schedules}
\end{table}

\begin{table}[h!]
    \centering
    \caption{Hyperparameter schedules for various FPMCs for the AFHQ-$64\times64$ dataset \citep{choi2020starganv2}, and EDM sampling schedule\citep{karras2022elucidating}}
    \begin{tabular}{l l | c c c c}
    \hline
    \multirow{2}{*}{Step} & \multirow{2}{*}{$t$} & PSPC-Square & LS & PSPC-Flex & Lukoianov \\
    & & $s(t)$ & $s(t)$ & $\tau(t)$ & $\tau$ \\
    \hline
    0 & 80 & 64 & 63 & 1.00 & 0.02 \\
    1 & 69.5 & 64 & 63 & 1.00 & 0.02 \\
    2 & 60.1 & 64 & 63 & 1.00 & 0.02 \\
    3 & 51.9 & 64 & 63 & 1.00 & 0.02 \\
    4 & 44.6 & 64 & 63 & 1.00 & 0.02 \\
    5 & 38.3 & 64 & 55 & 1.00 & 0.02 \\
    6 & 32.7 & 64 & 63 & 1.00 & 0.02 \\
    7 & 27.8 & 64 & 63 & 1.00 & 0.02 \\
    8 & 23.6 & 64 & 63 & 1.00 & 0.02 \\
    9 & 19.9 & 64 & 63 & 0.85 & 0.02 \\
    10 & 16.8 & 64 & 63 & 0.70 & 0.02 \\
    11 & 14.1 & 51 & 55 & 0.65 & 0.02 \\
    12 & 11.7 & 43 & 51 & 0.60 & 0.02 \\
    13 & 9.72 & 43 & 45 & 0.60 & 0.02 \\
    14 & 8.03 & 35 & 37 & 0.60 & 0.02 \\
    15 & 6.59 & 27 & 31 & 0.55 & 0.02 \\
    16 & 5.38 & 23 & 23 & 0.50 & 0.02 \\
    17 & 4.37 & 19 & 21 & 0.45 & 0.02 \\
    18 & 3.52 & 15 & 17 & 0.40 & 0.02 \\
    19 & 2.82 & 15 & 15 & 0.35 & 0.02 \\
    20 & 2.24 & 15 & 13 & 0.30 & 0.02 \\
    21 & 1.77 & 9 & 11 & 0.30 & 0.02 \\
    22 & 1.38 & 9 & 9 & 0.30 & 0.02 \\
    23 & 1.07 & 9 & 7 & 0.30 & 0.02 \\
    24 & 0.823 & 7 & 5 & 0.30 & 0.02 \\
    25 & 0.625 & 5 & 3 & 0.30 & 0.02 \\
    26 & 0.470 & 5 & 3 & 0.30 & 0.02 \\
    27 & 0.349 & 3 & 3 & 0.30 & 0.02 \\
    28 & 0.256 & 3 & 3 & 0.30 & 0.02 \\
    29 & 0.185 & 3 & 3 & 0.30 & 0.02 \\
    30 & 0.131 & 3 & 3 & 0.30 & 0.02 \\
    31 & 0.092 & 3 & 3 & 0.30 & 0.02 \\
    32 & 0.063 & 3 & 3 & 0.30 & 0.02 \\
    33 & 0.042 & 3 & 3 & 0.30 & 0.02 \\
    34 & 0.028 & 3 & 3 & 0.30 & 0.02 \\
    35 & 0.018 & 3 & 3 & 0.30 & 0.02 \\
    36 & 0.011 & 3 & 3 & 0.30 & 0.02 \\
    37 & 0.006 & 3 & 3 & 0.30 & 0.02 \\
    38 & 0.004 & 3 & 3 & 0.30 & 0.02 \\
    39 & 0.002 & 3 & 3 & 0.30 & 0.02 \\
    \hline
\end{tabular}
    \label{tab:afhq_schedules}
\end{table}
\subsection{Datasets}
\label{ap:datasets}

We evaluate our model on three datasets:
\begin{itemize}
    \item CIFAR-10 \citep{krizhevsky2009learning}, MIT License
    \item FFHQ-64$\times$64 \citep{karras2019style}, CC BY-NC-SA 4.0
    \item AFHQ-64$\times$64 \citep{choi2020starganv2}, CC BY-NC 4.0
\end{itemize}

For each dataset, we preprocess images such that they are floating point values on the range $[-1, 1]$

\subsection{Fine-tuning Details}
\label{ap:finetuning_details}
In this section, we detail the hyperparameters used to fine tune $\Qset$ and $\Rset$ both independently as presented in \cref{fig:finetuning}, as well as jointly for our methodology presented in \cref{tab:results}.

We paramaterize our soft $\Qset_\theta$ and $\Rset_\phi$ as $q_\ell = \exp(\theta_\ell)$ and $r_\ell = \exp(\phi_\ell)$ for $l = 1, \ldots, L$. To intialize training from binary $q$ or $r$ vectors, for all $\ell$, we set $\theta_\ell = \max(\log(q_\ell), 10^{-3})$ and $\phi_\ell = \max(\log(r_\ell), 10^{-3})$.

We minimize \cref{eq:obj} with respect to $\theta$, $\phi$ or both, depending on the experiment, with fixed $t$, and loss weight $\lambda(t)=1$. We use the AdamW optimizer \citep{loshchilov2017decoupled}, and use a constant learning rate of $0.05$ for all experiments with $\beta_1, \beta2= 0.9, 0.999$, and $\epsilon=1E-8$. We found weight decay of $\lambda=0.01$ was beneficial for $t\geq 1.92$ on CIFAR-10, $t\geq4.37$ on FFHQ and $t\geq8.03$ (steps 9,17, and 14 respectively), but use $\lambda=0$ otherwise. All models are trained with a batch size of 256, for 2000 steps or until convergence. We evaluate model performance on a validation set against the DDPM++ denoiser after each epoch, and select the best performing checkpoint across validation epochs. We use the test set of CIFAR-10, and a set of 10,000 synthetic diffusion model samples for FFHQ and AFHQ as a held-out validation set is not available for those datasets.

We run one optimization procedure per sampling timestep for the EDM sampling schedules of each dataset (values provided in \cref{tab:cifar_schedules,tab:ffhq_schedules,tab:afhq_schedules}). We use one Nvidia L40 48Gb GPU per training run. Total training time varies depending on dataset resolution, and the number of estimators in the FPMC, but a typical fine-tuning of PSPC-Flex on CIFAR-10 is completed in approximately 14 hours. 

Due to VRAM limitations, when training FFHQ we leverage Monte-Carlo score estimation techniques \citep{xu2023stable,niedoba2024nearest}, estimating the \cref{eq:fpmc_denoiser} using a random sample of $10,000$ $\bx \sim \nu(\bx)$ instead of iterating over the full dataset. We make this modification in the training step only, using the full average over $\text{supp}(\nu)$ for sampling and validation.

\subsection{Sampling Procedure}
\label{ap:sampling}
When generating images, we follow the deterministic sampling procedure of \cite{karras2022elucidating}, with diffusion parameters $\alpha(t)=1$, $\sigma(t)=t$. This corresponds to the PF-ODE
\begin{equation}
    d\bz = -t\nabla_\bz \log p_t(\bz) dt
\end{equation}
and a prior distribution $p_T(\bz)=\pi(\bz)=\mathcal{N}(0,T^2\mathbf{I})$, where $T=80$. We use the Heun second-order ODE solver, over $18$ values of $t$ for CIFAR-10 and $40$ values of $t$ for FFHQ$64\times64$, and AFHQ$64\times64$, for a total of 35 and 79 denoiser evaluations per sample respectively. We use a fixed set of initial 1000 $\bz$ for each dataset, shared among all denoisers.

% \subsection{Evaluation Metrics}

% We evaluate 
\section{Additional Results}
    \label{ap:additional}

\begin{figure}[h!]
    \centering
    \includegraphics[width=\linewidth]{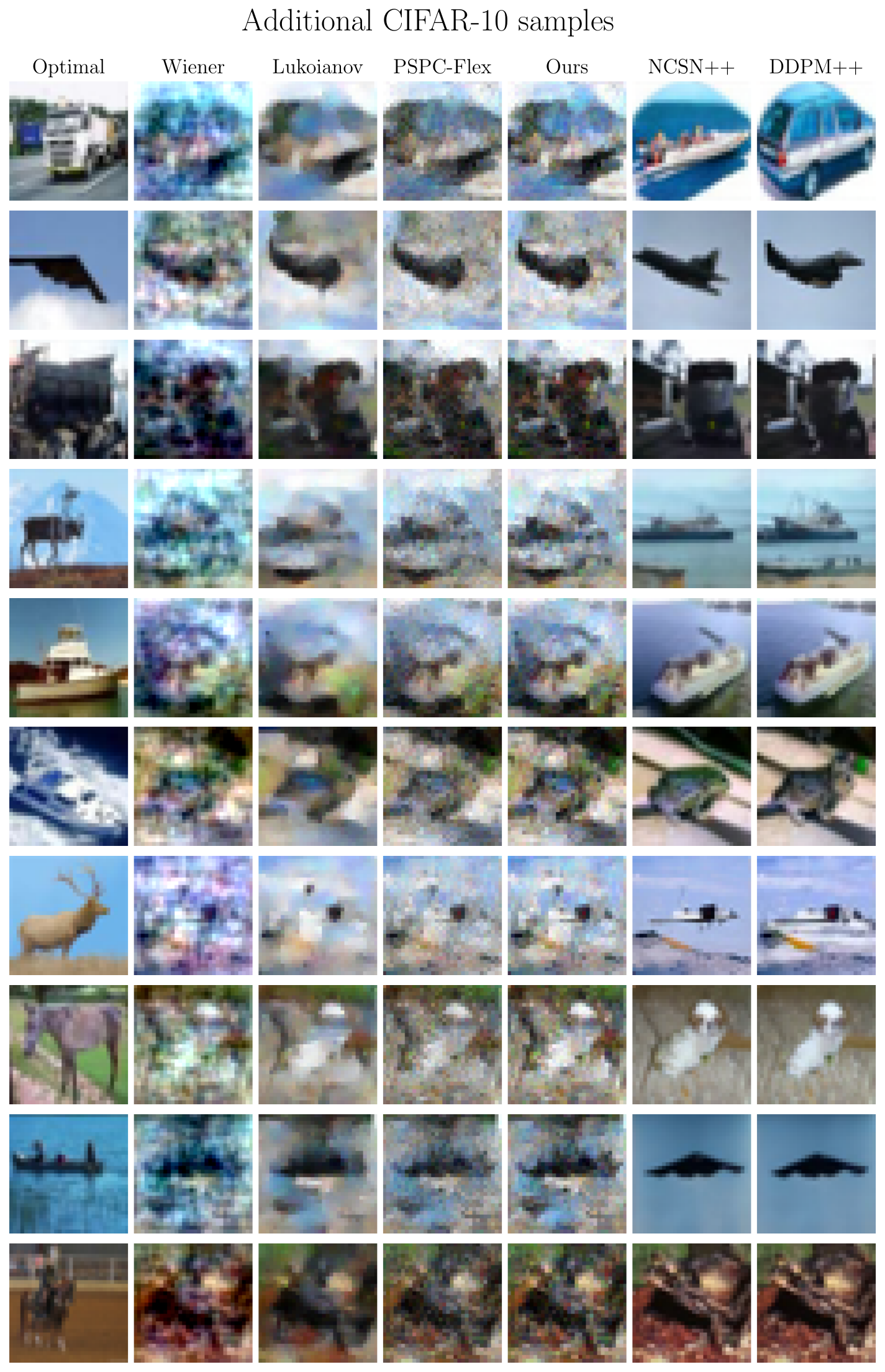}
    \caption{
         Comparison of additional samples generated by various denoisers with shared initial $\bz$ on CIFAR-10.
    }
    \label{fig:cifar_ap_samples}
\end{figure}

\begin{figure}[h!]
    \centering
    \includegraphics[width=\linewidth]{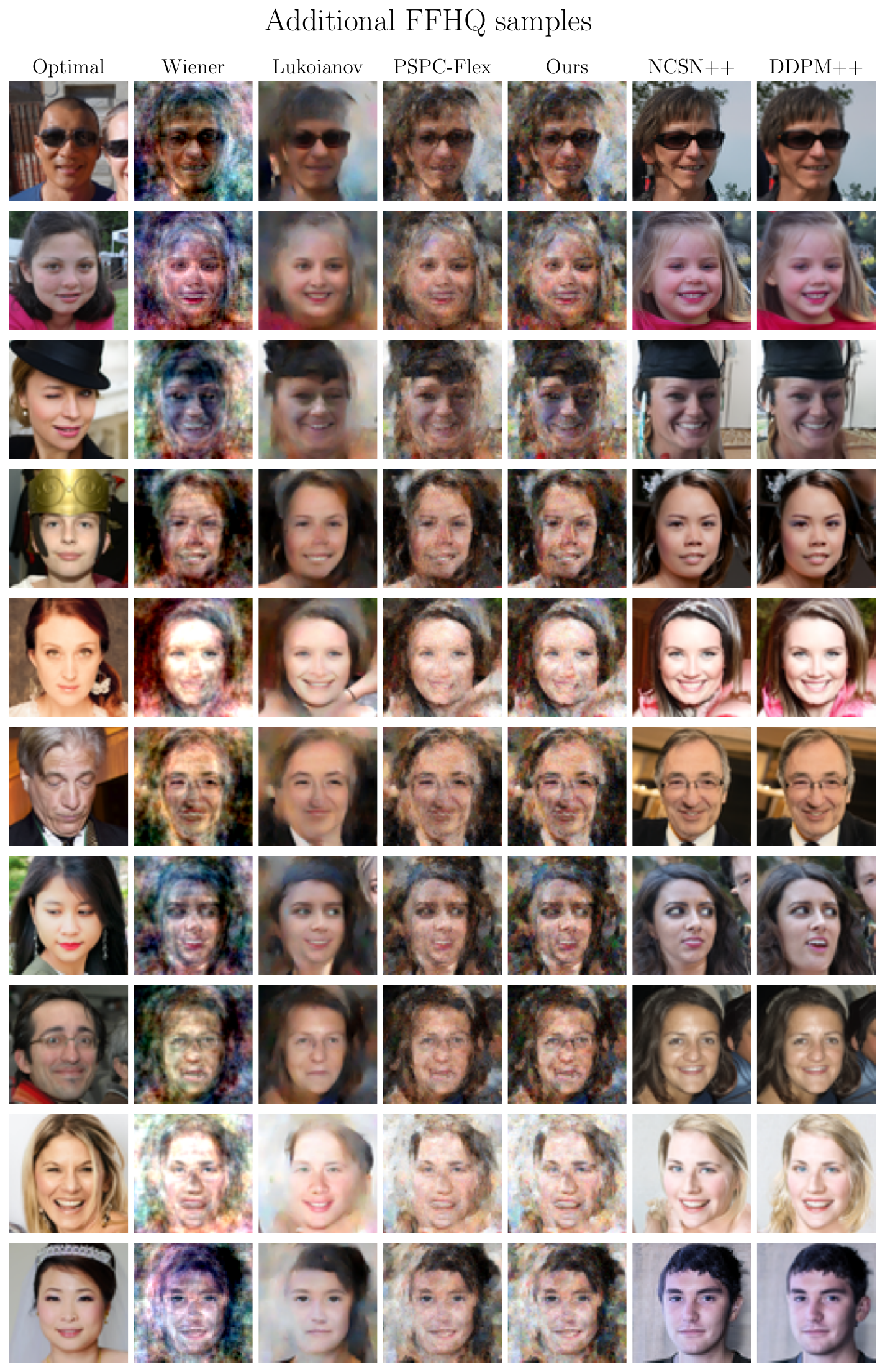}
    \caption{
         Comparison of additional samples generated by various denoisers with shared initial $\bz$ on FFHQ$64\times64$.
    }
    \label{fig:ffhq_ap_samples}
\end{figure}

\begin{figure}[h!]
    \centering
    \includegraphics[width=\linewidth]{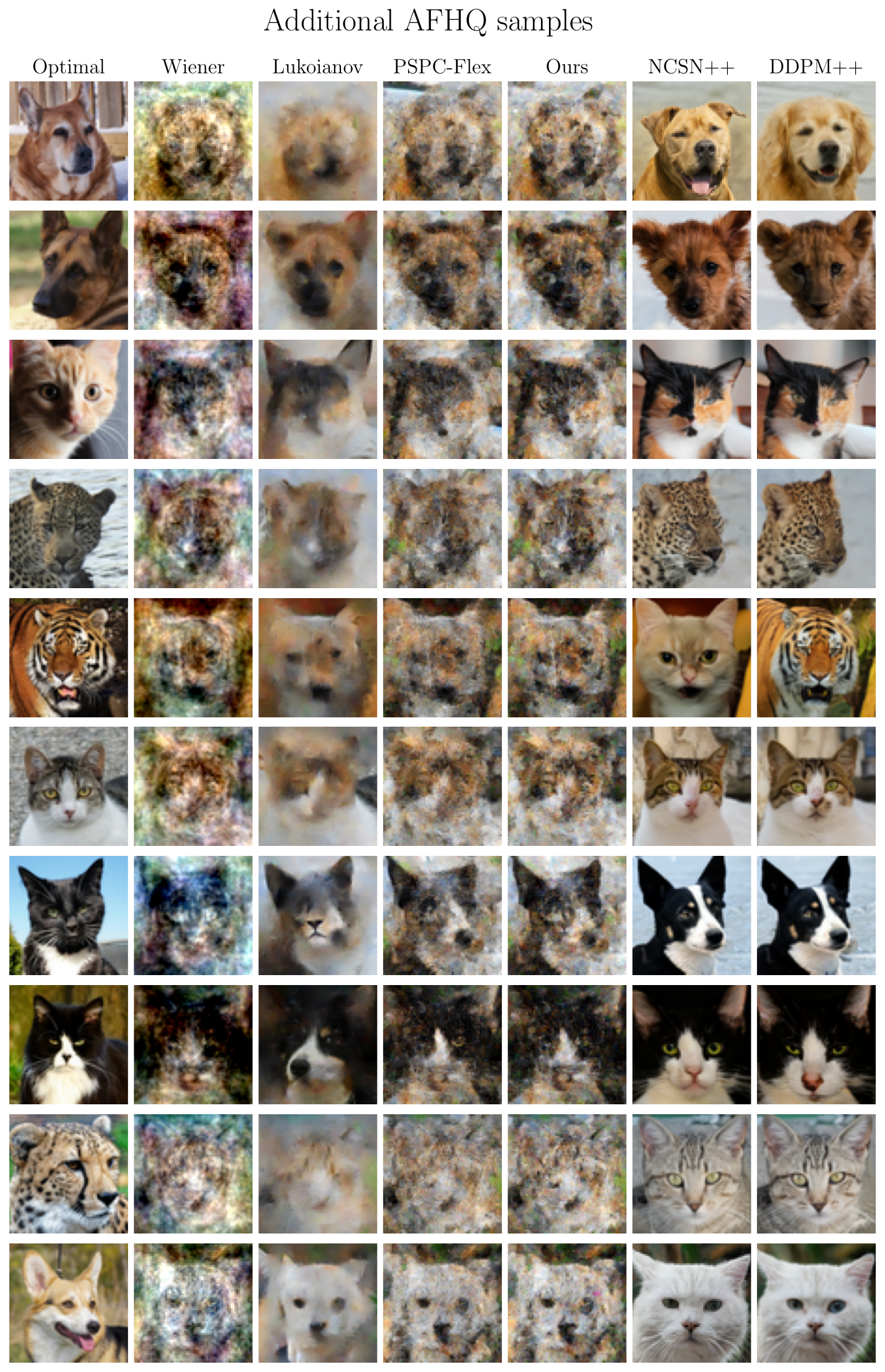}
    \caption{
         Comparison of additional samples generated by various denoisers with shared initial $\bz$ on AFHQ$64\times64$.}
    \label{fig:afhq_ap_samples}
\end{figure}

In this section, we provide additional samples drawn from each of the denoisers presented in \cref{fig:samples}. \Cref{fig:cifar_ap_samples,fig:ffhq_ap_samples,fig:afhq_ap_samples} show samples for CIFAR-10, FFHQ-$64\times64$ and AFHQ$64\times64$ respectively.

% \clearpage
% \section*{NeurIPS Paper Checklist}
%     \input{checklist}

% \input{instructions.tex}

% \newpage

\end{document}